\pdfoutput=1
\documentclass{article} %
\usepackage{iclr2027_conference,times}
\usepackage{microtype}
\usepackage{amsmath,amsfonts,bm}

\def\eqref#1{equation~\ref{#1}}

\def\1{\bm{1}}

\DeclareMathAlphabet{\mathsfit}{\encodingdefault}{\sfdefault}{m}{sl}
\SetMathAlphabet{\mathsfit}{bold}{\encodingdefault}{\sfdefault}{bx}{n}

\usepackage{hyperref}
\usepackage{url}
\usepackage{graphicx}
\usepackage[export]{adjustbox}  %
\usepackage{booktabs}
\usepackage{multirow}
\usepackage{makecell}
\usepackage{enumitem}
\usepackage{amsmath,amssymb,amsthm}
\usepackage{xcolor}
\usepackage[capitalize,noabbrev]{cleveref}
\newtheorem{proposition}{Proposition}

\newlength{\figoneH}\newlength{\figoneh}
\AtBeginDocument{\setlength{\figoneH}{0.3998\linewidth}%
  \setlength{\figoneh}{\dimexpr(\figoneH-4.4pt)/2\relax}}
\newcommand{\figoneleft}[1]{%
  \IfFileExists{#1}{\includegraphics[height=\figoneH]{#1}}%
    {\fbox{\parbox[c][\dimexpr\figoneH-2\fboxsep-2\fboxrule\relax][c]{\dimexpr0.6425\linewidth-2\fboxsep-2\fboxrule\relax}%
      {\centering\scriptsize\texttt{\detokenize{#1}}\\(17.21\,cm $\times$ 10.71\,cm)}}}}

\newlength{\figtwow}\newlength{\figtwoh}
\AtBeginDocument{\setlength{\figtwow}{\linewidth}%
  \setlength{\figtwoh}{0.4984\figtwow}}%
\newcommand{\figtwo}[1]{%
  \parbox[c][\figtwoh][c]{\figtwow}{\centering
    \IfFileExists{#1}%
      {\includegraphics[width=\figtwow,height=\figtwoh,keepaspectratio]{#1}}%
      {\fbox{\parbox[c][\dimexpr\figtwoh-2\fboxsep-2\fboxrule\relax][c]%
        {\dimexpr\figtwow-2\fboxsep-2\fboxrule\relax}%
        {\centering\small\texttt{\detokenize{#1}}}}}}}

\newcommand{\method}{\textsc{Decal}}
\usepackage{fancyvrb}
\usepackage{float}  %
\newcommand{\rev}[1]{#1}

\title{Quantizing Diffusion LLMs under Parallel\\ Decoding: Commitment Risk and the\\ Parallelism Tax}

\author{%
Huan Xu$^{1}$\thanks{Equal contribution.}\hphantom{$^{*}$}\quad
Maosen Zhao$^{1,2}$\footnotemark[1]\hphantom{$^{*}$}\quad
Pengtao Chen$^{1}$\quad
Tao Chen$^{1,2}$\thanks{Corresponding author.}\\
$^{1}$College of Future Information Technology, Fudan University, Shanghai, China\\
$^{2}$Shanghai Innovation Institute, Shanghai, China\\
\texttt{25213090239@m.fudan.edu.cn}, \texttt{eetchen@fudan.edu.cn}%
}

\iclrfinalcopy
\begin{document}

\maketitle
\lhead{Preprint}

\begin{abstract}
Diffusion large language models (dLLMs) are served with accelerated decoders that cache context and commit, in one step, every masked position whose confidence clears a threshold. Yet post-training quantization of dLLMs has so far been calibrated and evaluated under fixed-step samplers without caching. We study W4A4 quantization under the deployed decoder and find that the decoder determines both where quantization harms accuracy and what it costs. Irreversible errors enter chiefly through forced commitments, made when no position clears the threshold: they are a minority of commitments but carry most quantization-induced flips. Quantization also levies a \emph{parallelism tax}, extra forward passes per response: by compressing top logit gaps, it pushes positions below the threshold while leaving predictions largely intact. We propose \method{}, which reweights layer-wise calibration by a commitment-risk prior measured on the deployed decoder, and an argmax-preserving commit temperature that recovers the tax. On LLaDA and Dream, \method{} attains the best average accuracy among W4A4 methods, with lower run-to-run variance than the closest existing calibration method.
\end{abstract}

\section{Introduction}
\label{sec:intro}

Diffusion large language models (dLLMs) generate text by iteratively unmasking a response with bidirectional context \citep{austin2021d3pm,sahoo2024mdlm}, and open models such as LLaDA \citep{nie2025llada} and Dream \citep{ye2025dream} now rival autoregressive (AR) models of similar scale. Their speed in practice comes from accelerated decoders: Fast-dLLM \citep{wu2025fastdllm} combines block-wise approximate KV caching with confidence-thresholded parallel decoding, committing in one step every masked position whose confidence exceeds a threshold. Serving cost thus factors into the number of forward passes and the cost of each. Caching and parallel decoding reduce the former; low-bit post-training quantization (PTQ) \citep{frantar2023gptq}, for instance with rotations \citep[QuaRot;][]{ashkboos2024quarot} or learned affine transformations \citep[FlatQuant;][]{sun2025flatquant}, reduces the latter. Ideally, the two savings would compound.

\suppressfloats[t]
\begin{figure}[t]
  \centering
  \figoneleft{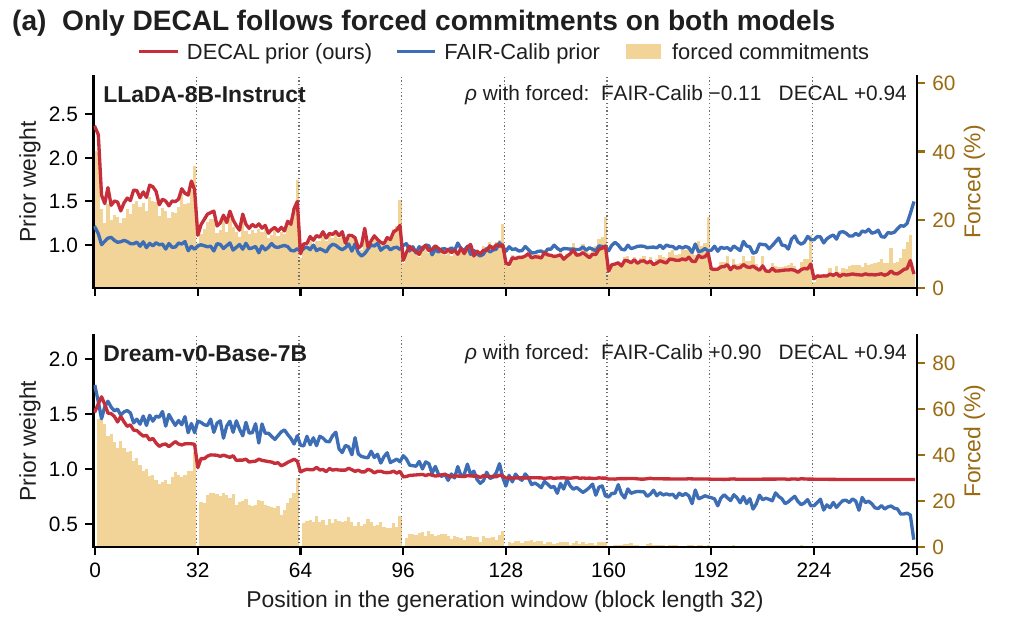}\hfill
  \parbox[b]{0.3366\linewidth}{%
    \includegraphics[height=\figoneh]{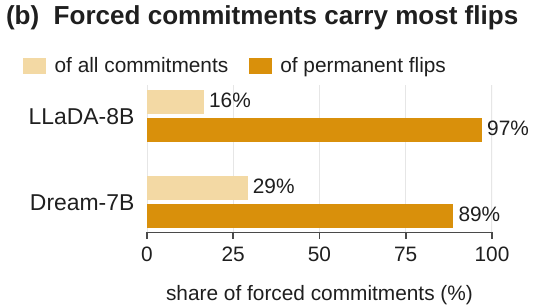}\\[4.4pt]
    \includegraphics[height=\figoneh]{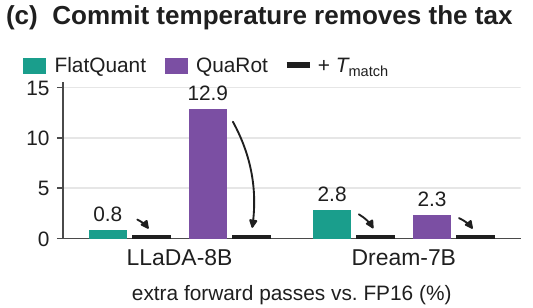}}
  \vspace{2pt}%
  \caption{\textbf{(a)}~Calibration priors of FAIR-Calib and \method{} over the 256 window positions (lines; mean 1), against the share of FP16 trajectories in which each position is a forced commitment under the deployed decoding rule (bars; GSM8K training prompts, no cache; Dream: returned output only); $\rho$: Spearman correlation. FAIR-Calib's policy-agnostic prior is uncorrelated with forced commitments on LLaDA and, on Dream, tracks their overall decline but not their jumps at block boundaries. \textbf{(b)}~Forced commitments are a minority of commitments but carry most permanent flips (GSM8K, W4A4 FlatQuant). \textbf{(c)}~Parallelism tax, the extra forward passes per response relative to FP16 (GSM8K), at $T=1$ (filled) and with the commit temperature at $T_{\mathrm{match}}$ (black line), counted as zero at or below FP16. Panels (b, c) use Fast-dLLM with dual cache and $\tau=0.9$.}
  \label{fig:obs}
\end{figure}

Existing dLLM quantization methods \citep{lin2025quantmeetsdllm,xu2025dllmquant,zhang2026quantdllm,yan2026starquant,huang2026faircalib}, however, are calibrated and evaluated apart from this decoder, under the model's native sampler, which commits a fixed number of positions per step without caching. The closest, FAIR-Calib \citep{huang2026faircalib}, reweights calibration by a position prior from a deliberately policy-agnostic probe; in our reproduction under Fast-dLLM, this prior is nearly uniform on LLaDA and does not follow where the decoder forces commitments (\cref{fig:obs}a). Yet under threshold decoding, the model's confidence, passed through the decoding rule, determines both how many forward passes a response takes and which decisions become irreversible.

We study W4A4 quantization directly under Fast-dLLM with dual caching, comparing each quantized model with a full-precision shadow evaluated on identical decoding states. Under the native sampler, the number of steps is fixed in advance and quantization can change only what is written; under threshold decoding, it also changes when positions are written and how many steps a response takes. This yields two observations with no counterpart under fixed-step samplers.

\textbf{Observation 1: irreversible errors enter through forced commitments.} A token is written either because its confidence exceeds the threshold (\emph{threshold commitment}) or because no position in the active block qualifies and the decoder must still write the most confident one (\emph{forced commitment}); either way, it is never revised. Forced commitments make up only 16--29\% of commitments yet carry 89--97\% of the \emph{permanent flips}, commitments that quantization changes and that survive into the returned output (\cref{fig:obs}b, \cref{sec:obs1}).

\textbf{Observation 2: quantization levies a parallelism tax.} Quantization deflates confidence while largely preserving what the model would write, so fewer positions clear the threshold at each step and each response takes more forward passes. We call this extra cost the \emph{parallelism tax}; it reaches 13\% on LLaDA and, as we show, can be recovered without retraining (\cref{fig:obs}c, \cref{sec:obs2}).

We therefore align both calibration and decoding with the deployed decoder. \method{} (\textbf{DE}coder-aligned \textbf{C}ommitment-risk c\textbf{AL}ibration) weights layer-wise calibration by a commitment-risk prior estimated from full-precision trajectories of the deployed decoder. A \emph{commit temperature} restores the confidence scale at the threshold test; since it preserves the argmax, it changes when positions are committed, never which token a position receives in a given state. \rev{The two act on different decisions: \method{} makes the fragile forced commitments less likely to flip, whereas the temperature releases only commitments just below the threshold, which rarely flip (\cref{app:division}).}

Our contributions are as follows:
\begin{itemize}
  \item To our knowledge, we are the first to systematically analyze PTQ for dLLMs under the cached, threshold-based parallel decoder used in deployment, identifying forced commitments, including a placement failure absent from AR decoding (writing the wrong position), as the entry point of irreversible errors, and characterizing the parallelism tax that quantization levies.\looseness=-1
  \item We propose \method{}, which scores each generation position by its flip probability, through the written token or the chosen position, times its reach into the returned output. Collected without any cache, the prior transfers to cached deployment, and its definition applies to both instruction-tuned and base models. \method{} attains the best average accuracy among W4A4 methods on both models, with and without cache, and, under cached decoding, a lower run-to-run variance than FAIR-Calib on every task.
  \item We introduce an argmax-preserving commit temperature, set by matching full-precision forward passes, that recovers the full tax of QuaRot on LLaDA without a detectable accuracy change and offers a controllable accuracy--efficiency trade-off beyond it.
\end{itemize}

\section{Related Work}
\label{sec:related}

\paragraph{Diffusion language models.}
Masked diffusion models, typically bidirectional Transformers \citep{vaswani2017attention}, generate text by iteratively denoising a fully masked sequence \citep{austin2021d3pm,lou2024sedd,sahoo2024mdlm,shi2024md4}. LLaDA \citep{nie2025llada} and Dream \citep{ye2025dream} scale them to billions of parameters. Because vanilla sampling recomputes full bidirectional attention at every step, a line of work accelerates dLLM inference with approximate KV caching \citep{ma2025dkvcache,liu2025dllmcache} and adaptive parallel decoding \citep{israel2025apd}; Fast-dLLM \citep{wu2025fastdllm} combines caching with threshold decoding and is the decoder we study. \rev{Polestar \citep{lee2026polestar} uses token representation drift both to refresh stale cache entries and to commit tokens below the confidence threshold when their drift shows a sharp event; under its drift-guided commitment, forced commitments still carry most quantization-induced flips, the parallelism tax persists, and both the prior of \method{} and the commit temperature keep their benefits (\cref{app:polestar}).\looseness=-1}

\paragraph{Post-training quantization of LLMs.}
Reconstruction-based PTQ fits quantization parameters layer by layer or block by block on a small calibration set \citep{nagel2020adaround,li2021brecq}. For LLMs, weight-only methods use second-order updates \citep{frantar2023gptq} or protect salient channels using activation statistics \citep{lin2024awq}, while weight--activation methods must handle activation outliers \citep{dettmers2022llmint8}, by smoothing \citep{xiao2023smoothquant}, rotation \citep{ashkboos2024quarot,liu2025spinquant}, or learned transformations and clipping \citep{shao2024omniquant,sun2025flatquant}.

\paragraph{Quantization of diffusion models and dLLMs.}
For image diffusion, calibration must cover the activations of all timesteps \citep{shang2023ptq4dm,li2023qdiffusion}. For dLLMs, \citet{lin2025quantmeetsdllm} show that AR PTQ methods transfer but W4A4 remains hard because of activation outliers. DLLMQuant \citep{xu2025dllmquant} and Quant-dLLM \citep{zhang2026quantdllm} calibrate across timesteps and mask ratios, STaR-Quant \citep{yan2026starquant} treats masked and unmasked token states separately, and FAIR-Calib \citep{huang2026faircalib} weights calibration positions by a prior probed from a full-precision teacher that commits one random position per step, with neither blocks nor a threshold. None is calibrated or evaluated under the cached threshold decoder used in deployment.

\paragraph{Temperature and confidence.}
Temperature scaling calibrates predicted probabilities post hoc \citep{guo2017calibration,minderer2021revisiting}; we use it to restore the scale the commit threshold reads.

\section{Observations}
\label{sec:obs}

We fix the decoder and the measurement, then establish the two observations our method builds on.

\subsection{Decoder and Measurement}
\label{sec:setting}

\paragraph{Threshold parallel decoding.}
Confidence-ordered parallel commitment dates back to Mask-Predict \citep{ghazvininejad2019maskpredict} and MaskGIT \citep{chang2022maskgit}; dLLM decoders apply it block by block, left to right \citep{arriola2025block}. A window of $K$ masked positions follows the prompt. At forward pass $t$, let $\mathcal{A}_t$ be the masked positions of the active block and, for $i\in\mathcal{A}_t$, let $z_t(i)$ be the logits, $p_t(i)=\max_v\mathrm{softmax}(z_t(i))_v$ the confidence, and $m_t(i)$ the gap between the two largest logits. Fast-dLLM \citep{wu2025fastdllm} commits
\begin{equation}
\mathcal{C}_t=
\begin{cases}
\{\,i\in\mathcal{A}_t:\ p_t(i)\ge\tau\,\} & \text{if non-empty (threshold commitments)},\\
\{c_t\},\ \ c_t=\arg\max_{i\in\mathcal{A}_t}p_t(i) & \text{otherwise (a forced commitment)},
\end{cases}
\label{eq:rule}
\end{equation}
and writes $\hat x_t(i)=\arg\max_v z_t(i)_v$ at every $i\in\mathcal{C}_t$; committed tokens are never revised. With dual caching, keys and values outside the active block are cached, refreshed at the first pass of each block (\emph{build step}) and read by the others (\emph{reuse steps}), a distinction specific to cached block decoding.\looseness=-1

\paragraph{Measurement.}
We let the quantized model decode and run its full-precision counterpart as a shadow on the same states, without letting it commit. A commitment is \emph{flipped} if the shadow would write a different token or, at a forced commitment, choose a different position; the flip is \emph{permanent} if the token is part of the returned output. We measure W4A4 LLaDA-8B-Instruct and Dream-v0-Base-7B on GSM8K with Fast-dLLM (dual cache, $\tau=0.9$).

\begin{figure}[t]
  \centering
  \includegraphics[width=\linewidth]{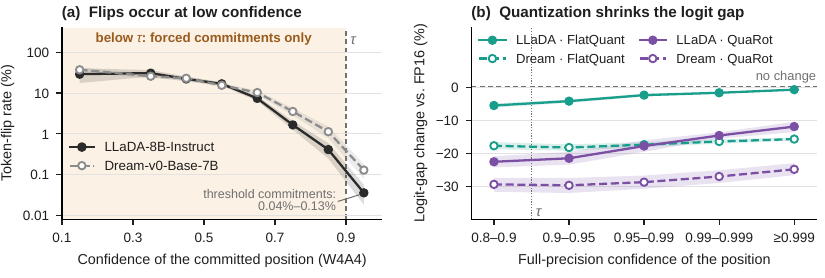}
  \caption{Evidence for the two observations (GSM8K, W4A4, against a full-precision shadow on the same states). \textbf{(a)}~Token-flip rate, the fraction of commitments whose token differs from the shadow's, by the quantized model's confidence at commitment (FlatQuant). \textbf{(b)}~Change of the gap between the two largest logits relative to full precision, grouped by full-precision confidence. Bands: 95\% confidence intervals.}
  \label{fig:mech}
\end{figure}

\subsection{Observation 1: Irreversible Errors Enter Through Forced Commitments}
\label{sec:obs1}

\paragraph{The threshold filters decisions by margin.}
A threshold commitment is made only when the top token leads its runner-up by a sizable logit margin, which a quantization perturbation can overturn only if it exceeds that margin: the threshold admits only robust decisions. The forced rule suspends this filter exactly when the model is uncertain, writing the most confident position at whatever margin it has; since nothing is revised, forced commitments are the channel through which fragile decisions become permanent. Accordingly, the flip rate falls steeply with the confidence at commitment and is up to three orders of magnitude higher below $\tau$, a range only forced commitments reach (\cref{fig:mech}a). With FlatQuant, forced commitments are 16.5\% of all commitments on LLaDA and 29.2\% on Dream, yet carry 97.1\% and 88.8\% of the permanent \rev{token} flips (\cref{fig:obs}b\rev{; 99.5\% and 96.7\% counting placement flips, \cref{app:forced}}): per commitment, 6.0\% and 9.8\% of them flip, against 0.04\% and 0.13\% of threshold commitments.

\paragraph{How forced commitments fail, and why only here.}
A forced commitment can flip its token, as in AR decoding, but also its position: when the most confident positions are nearly tied, quantization can change which one is written, a placement flip that AR decoding, with its fixed order, cannot make. Either error is permanent and becomes context for every later pass, so its damage grows with how long it is read before the returned output is complete. By contrast, a fixed-step sampler commits its $k$ most confident positions whatever their margins, spreading fragile commitments over the whole schedule, and a random-commit probe, as in FAIR-Calib, does not see them at all (\cref{fig:obs}a). Under threshold decoding, they are concentrated, can be located on full-precision trajectories before quantization, and differ in how far their errors reach; the prior of \cref{sec:decal} measures both.

\subsection{Observation 2: Quantization Levies a Parallelism Tax}
\label{sec:obs2}

\paragraph{An absolute test on a relative quantity.}
The threshold reads confidence as an absolute number, but confidence is set by the gaps between the top logit and its competitors: with two candidates, $p\ge\tau$ requires a lead of 2.2 nats at $\tau=0.9$ (\cref{sec:temp}). Anything that shrinks these gaps without changing the top token delays commitments while leaving every decision intact, and quantization does exactly this. It adds nearly independent noise to the logits, which barely changes their spread but pulls the top logit toward the rest and lifts the strongest of the many competitors, so the gaps near the peak shrink most: on LLaDA with QuaRot, the top-2 gap shrinks by 26\% while the standard deviation of the logits is unchanged (\cref{app:logits}), and across models and quantizers the shrinkage reaches 30\% at confident positions (\cref{fig:mech}b).

\paragraph{From compressed gaps to extra passes.}
Positions near $\tau$ then fall below it: fewer positions clear the threshold at each step, more steps end in a forced commitment, and each response needs more forward passes, up to 13\% more on LLaDA with QuaRot (\cref{fig:obs}c; both tasks in \cref{app:paired}). We call this cost the \emph{parallelism tax}. It compounds Observation~1, since the deflation that slows decoding also routes more decisions through the forced channel.

\paragraph{Deflation, not lost capability.}
The shadow of \cref{sec:setting} shows that the tax is a matter of scale. The width loss it measures step by step matches an independent end-to-end measurement (13.4\% against 14.6\% on 400 questions; \cref{app:paired}), so the extra passes come from fewer commitments per step, not from diverging trajectories; and where the shadow would commit but the quantized model withholds, the two predict the same token in 97.9\% of cases, against 68.3\% over all masked positions. Quantization perturbs predictions broadly but, where decisions are made, only lowers confidence, so rescaling confidence recovers the tax without touching written tokens (\cref{sec:temp}).

\begin{figure}[t]
  \centering
  \figtwo{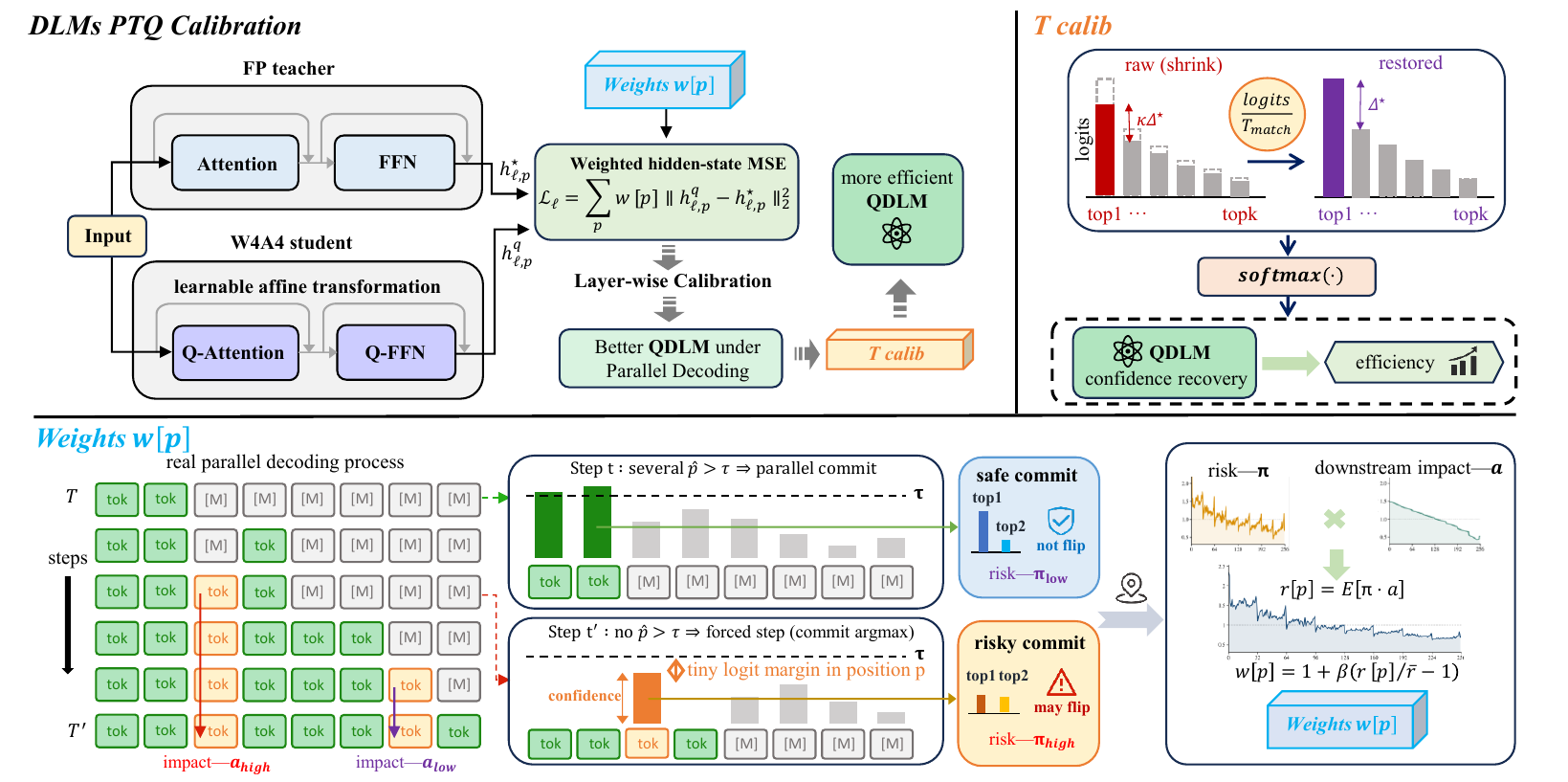}
  \caption{Overview of \method{} and the commit temperature. \textbf{Top left:} a W4A4 student is calibrated layer by layer against the full-precision teacher with a hidden-state MSE weighted by $w[p]$, and decodes with the commit temperature $T_{\mathrm{match}}$. \textbf{Bottom:} how $w[p]$ is obtained. Under the deployed parallel decoder, positions committed above $\tau$ are safe (low flip risk $\pi$), whereas forced commitments at small logit margins may flip (high $\pi$); positions committed early are read by more later passes (high reach $a$); $w[p]$ follows from $r[p]=\mathbb{E}[\pi\cdot a]$. \textbf{Top right:} dividing the logits by $T_{\mathrm{match}}$ restores the top-logit gaps that W4A4 compresses, and with them the full-precision commit rate.}
  \label{fig:main}
\end{figure}

\section{Method}
\label{sec:method}

\subsection{DECAL: Commitment-Risk Calibration}
\label{sec:decal}

\paragraph{Trajectories.}
\method{} decides \emph{where} the calibration budget is spent (\cref{fig:main}, bottom and top left). The full-precision model decodes $J$ unlabeled prompts under rule~(\ref{eq:rule}) with the deployed threshold, block length, and window, without cache. Position $p$ of trajectory $j$ is committed once, at pass $t_j(p)$, with logit gap $m_j(p)$; $L_j$ is the returned length after the stop rule (EOS or task stop sequence).

\paragraph{Flip probability.}
A commitment fails if the runner-up token overtakes the written one (\emph{token flip}) or, at a forced commitment, if the runner-up position overtakes the chosen one (\emph{placement flip}). Both are log-probability differences: the logit gap $m$ and the placement gap $g=\log p_t(c_t)-\max_{i\ne c_t}\log p_t(i)$. Modeling each difference's quantization perturbation as zero-mean Gaussian with scale $s$ gives flip probabilities $\Phi(-m/s)$ and $\Phi(-g/s)$; for independent failures,
\begin{equation}
\pi_j(p)=1-\Big(1-\Phi\big(-\tfrac{m_j(p)}{\tilde m}\big)\Big)\Big(1-\Phi\big(-\tfrac{g_j(p)}{\tilde m}\big)\Big),
\label{eq:flip}
\end{equation}
with $g_j(p)=+\infty$ except at forced commitments. With the quantizer not yet fixed, we set $s=\tilde m$, the median forced-commitment gap in returned outputs; only the ranking of $\pi$ matters, as~(\ref{eq:prior}) normalizes it. \rev{$\pi$ thus serves as a ranking score rather than a calibrated probability; \cref{app:flipmodel} checks this ranking and the assumptions of~(\ref{eq:flip}) against observed flips.}

\paragraph{Reach.}
A flipped token damages the returned output directly if returned, and indirectly through every later pass that reads it while returned tokens are still being written, up to $t^\star_j=\max_{p'<L_j}t_j(p')$:
\begin{equation}
a_j(p)=\sqrt{\mathbf{1}[p<L_j]+D_j(p)},\qquad D_j(p)=\big[\,t^\star_j-t_j(p)\,\big]_+ .
\label{eq:reach}
\end{equation}
The square root is the root-mean-square accumulation of approximately independent perturbations, a premise we verify in \cref{app:proof}. Reach gives no budget to a base model's discarded continuation, so one definition serves base and instruction-tuned models.

\paragraph{Prior and weights.}
Averaging over trajectories,
\begin{equation}
r(p)=\frac{1}{J}\sum_{j=1}^{J}\pi_j(p)\,a_j(p),\qquad
w(p)=1+\beta\Big(\frac{r(p)}{\bar r}-1\Big),\qquad
\beta=\frac{c}{\mathrm{CV}(r)} .
\label{eq:prior}
\end{equation}
As $\pi$ is the probability of an error and $a$ the damage it does, their product, not either factor alone, is an expected loss: if flipping and reaching are independent, $r(p)$ is the expected damage to the returned output, and it vanishes if either factor does. The affine map preserves the ranking and sets $\mathrm{CV}(w)=c$, a concentration budget trading focus against the effective sample size $K/(1+c^2)$.

\paragraph{Weighted layer-wise calibration.}
As in FAIR-Calib \citep{huang2026faircalib}, the prior reweights FlatQuant's layer-wise reconstruction \citep{sun2025flatquant} on generic text of length $T_{\mathrm c}$ whose last $K$ positions stand in for the window:
\begin{equation}
\min_{\Theta_\ell}\ \sum_{n=1}^{N}\sum_{i=0}^{T_{\mathrm c}-1}\omega_i\,\big\|h_\ell(\mathbf{x}_{n,i})-\hat h_\ell(\mathbf{x}_{n,i};\Theta_\ell)\big\|_2^2,
\label{eq:stage2}
\end{equation}
with $\omega_i=\lambda\,w(i{-}T_{\mathrm c}{+}K)$ on the last $K$ positions (window scale $\lambda$) and a context weight $\epsilon$ elsewhere; $h_\ell,\hat h_\ell$ are full-precision and quantized layer outputs, $\Theta_\ell$ the learnable transformations and clipping thresholds. As the cache approximates attention, not the decoding rule, the prior transfers to cached decoding: collecting it with dual cache instead, from a slightly smaller prompt set, leaves GSM8K accuracy on LLaDA unchanged (76.45\,$\pm$\,0.24 against 76.65\,$\pm$\,0.46, three runs each).

\subsection{Commit Temperature}
\label{sec:temp}

\paragraph{The threshold is a margin.}
The temperature sets \emph{when} positions commit (\cref{fig:main}, top right). With top token $v^\star$ and gaps $\Delta_v=z_{v^\star}-z_v\ge0$, the confidence and two-candidate test are
\begin{equation}
p=\Big[\,1+\sum\nolimits_{v\ne v^\star}e^{-\Delta_v}\Big]^{-1},
\qquad\qquad
p\ge\tau\iff\Delta\ge m_\tau:=\log\tfrac{\tau}{1-\tau}.
\label{eq:margin}
\end{equation}
The threshold thus fixes a margin in nats ($m_\tau=2.20$ at $\tau=0.9$) on the full-precision logit scale. If quantization compresses the top gaps as $\Delta^{\mathrm q}\approx\kappa\Delta$ with $\kappa<1$ (\cref{fig:mech}b), a position commits only when $\Delta\ge m_\tau/\kappa$: the effective threshold rises, which is the parallelism tax.

\paragraph{Definition.}
Because quantization compresses the logits whatever the decoder, the temperature applies to any confidence-thresholded decoder: rule~(\ref{eq:rule}) uses $p^{T}_t(i)=\max_v\mathrm{softmax}\big(z_t(i)/T\big)_v$ for the threshold test and for $c_t$, while the written token stays $\arg\max_v z_t(i)_v$. We study it in the deployed setting, cached block decoding, where we apply it on reuse steps only. Build steps keep $T=1$: a position withheld there disagrees with full precision more often than at reuse steps (3.2\% against 1.8\%), and a premature commitment there becomes context for the entire block \mbox{(\cref{app:design})}.

\begin{proposition}
\label{prop:temp}
For any decoding state and any $T\in(0,1]$ applied as above (proof in \cref{app:proof}):
\begin{enumerate}[label=(\roman*),leftmargin=2.2em,nosep]
\item \emph{content invariance:} every committed position receives the same token as at $T=1$;
\item \emph{monotone admission:} the number of committed positions is non-increasing in $T$, and a position admitted at $T$ but not at $T=1$ satisfies $T m_\tau\le\Delta<m_\tau$ (two candidates);
\item \emph{exact restoration:} if the logits contract uniformly, $z^{\mathrm q}=\kappa z+b\mathbf{1}$, then $p^{T}(z^{\mathrm q})=p(z)$ at $T=\kappa$.
\end{enumerate}
\end{proposition}
\rev{\Cref{prop:temp} is a per-state guarantee. No change to when positions are committed, not even lowering $\tau$, admits a trajectory-level guarantee, because an earlier commitment changes the context of every later step; we therefore measure the trajectory-level effects instead (\cref{app:ladder}).}

\paragraph{Calibration by matching the commit rate.}
Contraction is not uniform across positions and confidence levels (\cref{fig:mech}b), so no single $\kappa$ can be read off the logits. A threshold decoder, however, consumes confidence only through $\tau$-crossings, whose aggregate is the number of forward passes; we calibrate $T$ on exactly this statistic, the one for which $\tau$ was set:
\begin{equation}
T_{\mathrm{match}}=\sup\big\{\,T\in(0,1]:\ \bar N_{\mathrm q}(T)\le\bar N_{\mathrm{fp}}(1)\,\big\},
\label{eq:tmatch}
\end{equation}
where $\bar N_{M}(T)$ is the mean number of forward passes per response of model $M$ on calibration prompts. The criterion is \emph{label-free} (it never consults accuracy), \emph{consistent} (it returns $\kappa$ under uniform contraction, \cref{prop:temp}(iii)), \emph{self-disabling} (a quantizer without tax receives $T_{\mathrm{match}}=1$), and \emph{cheap}: by \cref{prop:temp}(ii), $\bar N_{\mathrm q}$ decreases with $T$, so two evaluations bracketing the target give $T_{\mathrm{match}}$ by linear interpolation. One temperature is calibrated per model--quantizer pair and reused on every task. Like temperature scaling \citep{guo2017calibration}, (\ref{eq:tmatch}) fits one scalar, but to the statistic its consumer uses: a classifier's temperature is fitted to the likelihood of labels, whereas a threshold decoder only asks whether confidence crosses $\tau$, so we fit the rate at which it does. With two candidates, a temperature commits when $\Delta\ge Tm_\tau$, as a lower threshold would, and at matched forward passes the two are indistinguishable (\cref{app:design}): (\ref{eq:tmatch}) adds not a new knob but a label-free way to set it.

\section{Experiments}
\label{sec:exp}
\suppressfloats[t]

\subsection{Setup}
\label{sec:exp-setup}

\paragraph{Models, decoding, and tasks.}
We quantize LLaDA-8B-Instruct \citep{nie2025llada} and Dream-v0-Base-7B \citep{ye2025dream} and decode with Fast-dLLM \citep{wu2025fastdllm} using dual cache, $\tau=0.9$, block length 32, and 256 generated tokens; we also report threshold decoding without cache. Following Fast-dLLM, we evaluate GSM8K \citep[5-shot, flexible-extract exact match;][]{cobbe2021gsm8k}, HumanEval \citep[0-shot, pass@1;][]{chen2021humaneval}, MBPP \citep[3-shot, pass@1;][]{austin2021mbpp}, and MATH \citep{hendrycks2021math} with the 4-shot Minerva prompt \citep{lewkowycz2022minerva}, 200 problems per subject, and an answer-equivalence check. These match Fast-dLLM's official evaluation; we report accuracy rather than perplexity, which scores teacher-forced text and never runs the decoder.\looseness=-1

\paragraph{Methods.}
All methods quantize weights and activations to 4 bits (W4A4). QuaRot \citep{ashkboos2024quarot} represents rotation-based PTQ, and FlatQuant \citep{sun2025flatquant} uses its default calibration with uniform weights. FAIR-Calib \citep{huang2026faircalib} and \method{} share FlatQuant's reconstruction on 128 WikiText-2 sequences \citep{merity2017wikitext} of length $T_{\mathrm c}=1{,}024$ whose last $K=256$ positions form the window, and differ only in the weights $\omega$ of~(\ref{eq:stage2}): the probed prior for FAIR-Calib (reproduced in our pipeline) and the commitment-risk prior~(\ref{eq:prior}) for \method{}, collected without cache from 5-shot GSM8K training prompts (about 640 per model; FAIR-Calib keeps its default 512, beyond which its own ablation shows no gain). This one-time cost is 2.8 GPU-hours on Dream (153 forward passes per prompt; LLaDA needs about half as many), fewer passes in total than FAIR-Calib's 512 probes of 256 steps each. FAIR-Calib, a prior over positions, is the dLLM calibration closest to ours; DLLMQuant and Quant-dLLM (2-bit, weight-only) instead change the calibration data or the quantizer, and are orthogonal to the prior. \method{} uses $(\epsilon,c)=(0.01,0.30)$ with FAIR-Calib's window scale on LLaDA, and $(1,0.15)$ with unit-mean window weights on Dream. \rev{The concentration $c$ is set once per model on the GSM8K test set and then frozen for every other task and every calibration run; it only sets how strongly the budget is concentrated, while the ranking of positions comes from the decoder. Once $c$ reaches a moderate concentration, the gain stays stable up to near its largest admissible value (\cref{app:csens}).} The commit temperature uses GSM8K forward-pass counts only (full runs for LLaDA, 50 prompts for Dream). Under dual cache, we repeat calibration three times for the FlatQuant-based methods and report the mean and standard deviation; FP16 and QuaRot are single runs. Decoding without cache is substantially slower, and we run each configuration once. All accuracy tables use $T=1$; we evaluate the commit temperature in the deployed setting, with dual cache (\cref{sec:exp-temp}).

\subsection{Main Results}
\label{sec:exp-main}

\begin{table}[!t]
\centering\small
\caption{W4A4 accuracy (\%) under the deployed decoder (Fast-dLLM, dual cache, $\tau=0.9$); FlatQuant-based methods: mean$\,\pm\,$std over three calibration runs; best W4A4 result in bold.}
\label{tab:main}
\setlength{\tabcolsep}{4.2pt}
\begin{tabular}{llccccc}
\toprule
Model & Method & GSM8K & HumanEval & MBPP & MATH & Avg. \\
\midrule
\multirow{5}{*}{\makecell{LLaDA-8B\\Instruct}}
 & FP16       & 78.70 & 36.59 & 36.40 & 30.79 & 45.62 \\
 & QuaRot     & 74.30 & 33.54 & 32.20 & 26.86 & 41.73 \\
 & FlatQuant  & 75.44\,{\scriptsize$\pm$1.07} & 35.57\,{\scriptsize$\pm$3.97} & 35.80\,{\scriptsize$\pm$2.43} & 28.21\,{\scriptsize$\pm$0.22} & 43.76 \\
 & FAIR-Calib & 75.84\,{\scriptsize$\pm$0.88} & 35.77\,{\scriptsize$\pm$1.86} & \textbf{36.12}\,{\scriptsize$\pm$1.41} & 28.43\,{\scriptsize$\pm$2.29} & 44.04 \\
 & \method{}  & \textbf{76.65}\,{\scriptsize$\pm$0.46} & \textbf{36.21}\,{\scriptsize$\pm$0.93} & 36.08\,{\scriptsize$\pm$1.06} & \textbf{28.67}\,{\scriptsize$\pm$0.24} & \textbf{44.40} \\
\midrule
\multirow{5}{*}{\makecell{Dream-v0\\Base-7B}}
 & FP16       & 73.16 & 53.05 & 51.00 & 30.79 & 52.00 \\
 & QuaRot     & 53.30 & 23.78 & 37.80 & 17.64 & 33.13 \\
 & FlatQuant  & 66.34\,{\scriptsize$\pm$1.05} & 40.85\,{\scriptsize$\pm$3.66} & \textbf{47.40}\,{\scriptsize$\pm$1.39} & 23.57\,{\scriptsize$\pm$0.97} & 44.54 \\
 & FAIR-Calib & 66.47\,{\scriptsize$\pm$1.47} & 40.88\,{\scriptsize$\pm$4.25} & 46.33\,{\scriptsize$\pm$2.19} & 23.43\,{\scriptsize$\pm$0.86} & 44.28 \\
 & \method{}  & \textbf{67.27}\,{\scriptsize$\pm$1.08} & \textbf{41.81}\,{\scriptsize$\pm$2.11} & 46.20\,{\scriptsize$\pm$1.78} & \textbf{24.60}\,{\scriptsize$\pm$0.72} & \textbf{44.97} \\
\bottomrule
\end{tabular}
\end{table}

\Cref{tab:main} reports the deployed setting. Rotation alone is not enough at W4A4: QuaRot loses 3.9 average points on LLaDA and 18.9 on the base model Dream, whereas the FlatQuant-based methods retain most of the full-precision accuracy. Among them, \method{} has the best average on both models and the best GSM8K, HumanEval, and MATH accuracy; on MBPP, the three lie within one standard deviation of one another. Although the prior is collected on GSM8K prompts alone, its gains carry over to HumanEval, as it encodes where the decoder commits fragile tokens, a property of the decoding schedule rather than of the domain. \method{} is also more stable than FAIR-Calib: its standard deviation across runs is lower on all eight model--task pairs, often by about half (0.46 against 0.88 on LLaDA GSM8K, 2.11 against 4.25 on Dream HumanEval, 0.24 against 2.29 on LLaDA MATH). Without cache (\cref{tab:nocache}), the regime in which its prior is collected, \method{} again has the best average on both models, with the widest margin on Dream (47.52 against 44.98 for FAIR-Calib), so its gains do not hinge on the cache. \rev{On the block-diffusion model Fast-dLLM v2 \citep{wu2026fastdllmv2}, forced commitments are nearly as common as on LLaDA, quantization levies a parallelism tax, and \method{} again improves W4A4 accuracy on its 7B model (\cref{app:v2}).}

\begin{table}[!t]
\centering\small
\caption{W4A4 accuracy (\%) with threshold decoding without cache ($\tau=0.9$; one run per configuration).}
\label{tab:nocache}
\footnotesize
\setlength{\tabcolsep}{3.0pt}
\begin{tabular}{l ccccc ccccc}
\toprule
 & \multicolumn{5}{c}{LLaDA-8B-Instruct} & \multicolumn{5}{c}{Dream-v0-Base-7B} \\
\cmidrule(lr){2-6}\cmidrule(lr){7-11}
Method & GSM8K & HumanEval & MBPP & MATH & Avg. & GSM8K & HumanEval & MBPP & MATH & Avg. \\
\midrule
FP16       & 78.09 & 42.07 & 40.20 & 32.00 & 48.09 & 72.40 & 49.39 & 53.80 & 31.86 & 51.86 \\
QuaRot     & 73.46 & 34.76 & 38.40 & 28.21 & 43.71 & 54.89 & 29.88 & 39.20 & 18.86 & 35.71 \\
FlatQuant  & 76.35 & 33.54 & \textbf{40.60} & 29.64 & 45.03 & 66.41 & 35.37 & 47.20 & 24.50 & 43.37 \\
FAIR-Calib & 76.27 & 38.90 & 40.21 & 29.57 & 46.24 & 64.06 & \textbf{43.90} & 46.40 & 25.57 & 44.98 \\
\method{}  & \textbf{76.95} & \textbf{39.02} & 39.60 & \textbf{29.80} & \textbf{46.34} & \textbf{68.16} & 43.29 & \textbf{52.00} & \textbf{26.64} & \textbf{47.52} \\
\bottomrule
\end{tabular}
\end{table}

\begin{table}[!t]
\centering\small
\caption{Components of the prior under dual cache (mean of two calibration runs). Uniform sets $w\equiv1$, single-factor variants set the other factor of~(\ref{eq:prior}) to one\rev{; shuffled permutes the weights of \method{} over the 256 window positions}; $c$ and reconstruction are shared.}
\label{tab:ablation}
\setlength{\tabcolsep}{5pt}
\begin{tabular}{l ccc ccc}
\toprule
 & \multicolumn{3}{c}{LLaDA-8B-Instruct} & \multicolumn{3}{c}{Dream-v0-Base-7B} \\
\cmidrule(lr){2-4}\cmidrule(lr){5-7}
Prior & GSM8K & MATH & Avg. & GSM8K & MATH & Avg. \\
\midrule
Uniform ($w\equiv1$)                 & 74.91 & 28.22 & 51.57 & 66.00 & 23.11 & 44.56 \\
\rev{Shuffled \method{} weights} & \rev{74.68} & \rev{28.47} & \rev{51.58} & \rev{66.46} & \rev{24.79} & \rev{45.63} \\
Flip probability only ($a\equiv1$)   & 76.04 & \textbf{29.32} & 52.68 & \textbf{67.78} & 24.81 & 46.30 \\
Reach only ($\pi\equiv1$)            & 76.17 & 29.11 & 52.64 & 65.97 & \textbf{25.89} & 45.93 \\
\method{} ($\pi\cdot a$)             & \textbf{76.65} & 28.85 & \textbf{52.75} & 67.74 & 24.90 & \textbf{46.32} \\
\bottomrule
\end{tabular}
\end{table}

\subsection{Ablation: Components of the Prior}
\label{sec:exp-ablation}

\Cref{tab:ablation} isolates the two factors of~(\ref{eq:prior}). Every risk-aware prior improves on uniform weights, by 1.1--1.2 average points on LLaDA and 1.4--1.8 on Dream: where the calibration budget goes matters. \rev{The shuffled control, with the same weights at random positions, gains nothing on LLaDA and least on Dream, so the gain comes from where the weights are placed, not from non-uniform weighting as such.} How the factors share it depends on the model and the task, as follows from what each measures: $\pi$ marks \emph{locally fragile} commitments, $a$ commitments whose errors \emph{propagate} through later reads.\looseness=-1

\paragraph{Task: local versus propagating errors.}
GSM8K solutions are short, so correctness hinges on a few decisive tokens. MATH solutions are long derivations in which an early error propagates into everything written after it. On Dream, the factors separate: $\pi$ alone carries the GSM8K gain (+1.8 points, against none for $a$ alone), whereas $a$ alone gives the largest MATH gain (+2.8 against +1.7).

\paragraph{Model: why the separation appears on Dream.}
An instruction-tuned model stops on its own, so its commitments, fragile or not, lie inside the returned answer: on LLaDA the factors largely agree, and either alone recovers most of the gain (+0.9 to +1.3 points). A base model writes past its answer, so part of its fragile commitments falls in a discarded continuation while reach concentrates on the answer: on Dream the factors single out different positions, and the task decides which matters. \rev{\Cref{tab:ablation} does not separate the product from the stronger single factor, which lies within 0.1 average points of it on both models. We keep the product as} the robust choice: \method{} need not know which failure mode a task exhibits, has the best average on both models, and stays within one point of the best variant everywhere, whereas each factor alone falls short somewhere.

\subsection{Calibrating the Commit Temperature}
\label{sec:exp-temp}

\begin{figure}[!ht]
  \centering
  \includegraphics[width=0.98\linewidth]{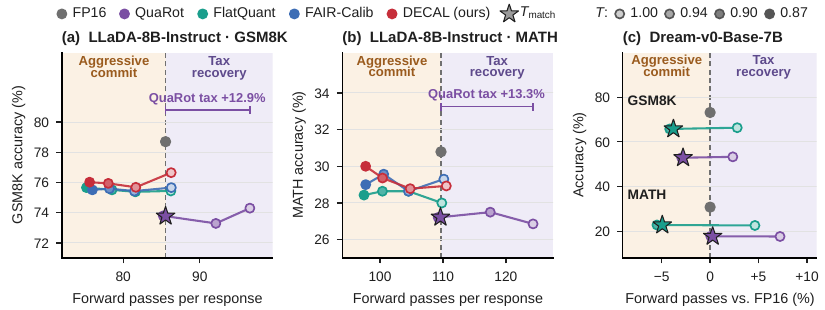}
  \caption{Commit temperature: accuracy against forward passes (\textbf{a,\,b}: LLaDA-8B-Instruct; \textbf{c}: Dream-v0-Base-7B, relative to FP16). Colors: methods; darker shades: lower $T$; stars: $T_{\mathrm{match}}$~(\ref{eq:tmatch}), interpolated (omitted where it is ${\approx}\,0.99$, i.e., at $T=1$); dashed line: FP16 at $T=1$.}
  \label{fig:temp}
\end{figure}

\paragraph{Calibration recovers the tax where it exists, and only there.}
We evaluate each LLaDA configuration at $T\in\{1,0.94,0.90,0.87\}$ and set $T_{\mathrm{match}}$ by~(\ref{eq:tmatch}) from GSM8K forward-pass counts alone, holding out MATH. QuaRot pays a 13\% tax on both tasks (\cref{fig:temp}a,\,b). At its $T_{\mathrm{match}}=0.873$, it returns to the full-precision forward-pass count, while accuracy moves by less than 0.6 points, within single-run noise ($\pm2.4$); solving~(\ref{eq:tmatch}) on MATH alone gives 0.874, so the GSM8K-calibrated temperature carries over to the held-out task. FlatQuant, FAIR-Calib, and \method{} pay under 1\%, receive $T_{\mathrm{match}}\approx0.99$, and are left essentially unchanged.

\paragraph{Below $T_{\mathrm{match}}$, it is an explicit trade-off.}
At $T=0.87$, these three models need about 11\% fewer forward passes than FP16, with accuracy within about one point of $T=1$ across the ladder.

\paragraph{It transfers across models.}
On Dream, $T_{\mathrm{match}}$ calibrated on only 50 GSM8K prompts removes taxes of 2--7\% on both tasks with accuracy changes below 0.6 points (\cref{fig:temp}c); the small calibration set slightly overestimates the tax, leaving up to 5\% fewer forward passes than FP16. Lowering $T$ only reorders commitments: at each question's first early release, the written token is the one the $T=1$ schedule eventually writes in at least 98.7\% of questions (\cref{app:ladder}).

\section{Conclusion}
\label{sec:conclusion}

We studied W4A4 quantization of dLLMs under the decoder they are served with. Under cached threshold decoding, quantization errors become irreversible chiefly through forced commitments, and quantization levies a parallelism tax by deflating confidence rather than changing predictions. \method{} weights calibration by commitment risk, and a commit temperature calibrated on forward passes alone recovers the tax without retraining or labels. On LLaDA and Dream, \method{} attains the best average accuracy among W4A4 methods with lower run-to-run variance than FAIR-Calib, and the temperature restores full-precision decoding cost. The prior is tied to the deployed block length and threshold, and we have studied one decoder family at one bit width; extending commitment-aware calibration to other decoders and lower bit widths is a natural next step.

\subsection*{AI use statement}
In this work, we used generative AI tools to assist with writing, including drafting, editing, and polishing the text, and with parts of the code, including analysis and plotting scripts. We have reviewed all AI-assisted work: the authors checked and revised all text, and verified and tested all AI-assisted code; every reported number comes from our own experiments. We take responsibility for the final content of this work, including text, claims or artifacts produced with the aid of generative AI.\looseness=-1

\subsection*{Reproducibility statement}
\Cref{sec:exp-setup} lists the models, decoder settings, tasks, calibration data, and hyperparameters of every method; \cref{sec:decal,sec:temp} define the prior, the weighted reconstruction, and the calibration of the commit temperature; \cref{app:paired,app:logits,app:ladder,app:strata,app:design} describe the measurement protocols behind the observations and the temperature study; and \cref{app:proof} proves \cref{prop:temp}\rev{; \cref{app:forced,app:division,app:flipmodel,app:polestar,app:v2,app:csens} report further analyses}. We will release the code for collecting the prior, the weighted calibration, the commit temperature, and the evaluation.

\subsection*{Acknowledgments}
This work is supported by National Key R\&D Program of China (No.~2026YFE0101200), National Natural Science Foundation of China (No.~62671175), and Shanghai General AI Foundation Models Program (No.~2025SHZDZX026D38). The computations in this research were performed using the CFFF platform of Fudan University.

\clearpage
\bibliography{references}

\begin{thebibliography}{43}
\providecommand{\natexlab}[1]{#1}
\providecommand{\url}[1]{\texttt{#1}}
\expandafter\ifx\csname urlstyle\endcsname\relax
  \providecommand{\doi}[1]{doi: #1}\else
  \providecommand{\doi}{doi: \begingroup \urlstyle{rm}\Url}\fi

\bibitem[Arriola et~al.(2025)Arriola, Gokaslan, Chiu, Yang, Qi, Han, Sahoo, and
  Kuleshov]{arriola2025block}
Marianne Arriola, Aaron Gokaslan, Justin~T. Chiu, Zhihan Yang, Zhixuan Qi,
  Jiaqi Han, Subham~Sekhar Sahoo, and Volodymyr Kuleshov.
\newblock Block diffusion: Interpolating between autoregressive and diffusion
  language models.
\newblock In \emph{International Conference on Learning Representations}, 2025.

\bibitem[Ashkboos et~al.(2024)Ashkboos, Mohtashami, Croci, Li, Cameron, Jaggi,
  Alistarh, Hoefler, and Hensman]{ashkboos2024quarot}
Saleh Ashkboos, Amirkeivan Mohtashami, Maximilian~L. Croci, Bo~Li, Pashmina
  Cameron, Martin Jaggi, Dan Alistarh, Torsten Hoefler, and James Hensman.
\newblock {QuaRot}: Outlier-free 4-bit inference in rotated {LLMs}.
\newblock In \emph{Advances in Neural Information Processing Systems},
  volume~37, 2024.

\bibitem[Austin et~al.(2021{\natexlab{a}})Austin, Johnson, Ho, Tarlow, and
  van~den Berg]{austin2021d3pm}
Jacob Austin, Daniel~D. Johnson, Jonathan Ho, Daniel Tarlow, and Rianne van~den
  Berg.
\newblock Structured denoising diffusion models in discrete state-spaces.
\newblock In \emph{Advances in Neural Information Processing Systems},
  volume~34, 2021{\natexlab{a}}.

\bibitem[Austin et~al.(2021{\natexlab{b}})Austin, Odena, Nye, Bosma,
  Michalewski, Dohan, Jiang, Cai, Terry, Le, and Sutton]{austin2021mbpp}
Jacob Austin, Augustus Odena, Maxwell Nye, Maarten Bosma, Henryk Michalewski,
  David Dohan, Ellen Jiang, Carrie Cai, Michael Terry, Quoc Le, and Charles
  Sutton.
\newblock Program synthesis with large language models.
\newblock \emph{arXiv preprint arXiv:2108.07732}, 2021{\natexlab{b}}.

\bibitem[Bie et~al.(2025)Bie, Cao, Chen, Du, et~al.]{bie2025llada2}
Tiwei Bie, Maosong Cao, Kun Chen, Lun Du, et~al.
\newblock {LLaDA2.0}: Scaling up diffusion language models to 100{B}.
\newblock \emph{arXiv preprint arXiv:2512.15745}, 2025.

\bibitem[Chang et~al.(2022)Chang, Zhang, Jiang, Liu, and
  Freeman]{chang2022maskgit}
Huiwen Chang, Han Zhang, Lu~Jiang, Ce~Liu, and William~T. Freeman.
\newblock {MaskGIT}: Masked generative image transformer.
\newblock In \emph{Proceedings of the IEEE/CVF Conference on Computer Vision
  and Pattern Recognition (CVPR)}, 2022.

\bibitem[Chen et~al.(2021)Chen, Tworek, Jun, Yuan, et~al.]{chen2021humaneval}
Mark Chen, Jerry Tworek, Heewoo Jun, Qiming Yuan, et~al.
\newblock Evaluating large language models trained on code.
\newblock \emph{arXiv preprint arXiv:2107.03374}, 2021.

\bibitem[Cobbe et~al.(2021)Cobbe, Kosaraju, Bavarian, Chen, Jun, Kaiser,
  Plappert, Tworek, Hilton, Nakano, Hesse, and Schulman]{cobbe2021gsm8k}
Karl Cobbe, Vineet Kosaraju, Mohammad Bavarian, Mark Chen, Heewoo Jun, Lukasz
  Kaiser, Matthias Plappert, Jerry Tworek, Jacob Hilton, Reiichiro Nakano,
  Christopher Hesse, and John Schulman.
\newblock Training verifiers to solve math word problems.
\newblock \emph{arXiv preprint arXiv:2110.14168}, 2021.

\bibitem[Dettmers et~al.(2022)Dettmers, Lewis, Belkada, and
  Zettlemoyer]{dettmers2022llmint8}
Tim Dettmers, Mike Lewis, Younes Belkada, and Luke Zettlemoyer.
\newblock {LLM.int8()}: 8-bit matrix multiplication for transformers at scale.
\newblock In \emph{Advances in Neural Information Processing Systems},
  volume~35, 2022.

\bibitem[Frantar et~al.(2023)Frantar, Ashkboos, Hoefler, and
  Alistarh]{frantar2023gptq}
Elias Frantar, Saleh Ashkboos, Torsten Hoefler, and Dan Alistarh.
\newblock {GPTQ}: Accurate post-training quantization for generative
  pre-trained transformers.
\newblock In \emph{International Conference on Learning Representations}, 2023.

\bibitem[Ghazvininejad et~al.(2019)Ghazvininejad, Levy, Liu, and
  Zettlemoyer]{ghazvininejad2019maskpredict}
Marjan Ghazvininejad, Omer Levy, Yinhan Liu, and Luke Zettlemoyer.
\newblock {Mask-Predict}: Parallel decoding of conditional masked language
  models.
\newblock In \emph{Proceedings of the 2019 Conference on Empirical Methods in
  Natural Language Processing and the 9th International Joint Conference on
  Natural Language Processing (EMNLP-IJCNLP)}, 2019.

\bibitem[Guo et~al.(2017)Guo, Pleiss, Sun, and Weinberger]{guo2017calibration}
Chuan Guo, Geoff Pleiss, Yu~Sun, and Kilian~Q. Weinberger.
\newblock On calibration of modern neural networks.
\newblock In \emph{International Conference on Machine Learning}, 2017.

\bibitem[Hendrycks et~al.(2021)Hendrycks, Burns, Kadavath, Arora, Basart, Tang,
  Song, and Steinhardt]{hendrycks2021math}
Dan Hendrycks, Collin Burns, Saurav Kadavath, Akul Arora, Steven Basart, Eric
  Tang, Dawn Song, and Jacob Steinhardt.
\newblock Measuring mathematical problem solving with the {MATH} dataset.
\newblock In \emph{Advances in Neural Information Processing Systems, Datasets
  and Benchmarks Track}, 2021.

\bibitem[Huang et~al.(2026)Huang, Yang, Xu, Liu, Guo, Fu, Zhou, and
  Zhang]{huang2026faircalib}
Haoyu Huang, Linlin Yang, Sheng Xu, Boyu Liu, Guodong Guo, Zhongqian Fu, Hang
  Zhou, and Baochang Zhang.
\newblock {FAIR-Calib}: Frontier-aware instability-reweighted calibration for
  post-training quantization of diffusion large language models.
\newblock In \emph{International Conference on Machine Learning}, 2026.

\bibitem[Israel et~al.(2025)Israel, Van~den Broeck, and Grover]{israel2025apd}
Daniel Israel, Guy Van~den Broeck, and Aditya Grover.
\newblock Accelerating diffusion {LLMs} via adaptive parallel decoding.
\newblock In \emph{Advances in Neural Information Processing Systems},
  volume~38, 2025.

\bibitem[Lee et~al.(2026)Lee, Ramachandran, Kundu, and
  Krishna]{lee2026polestar}
Mingyu Lee, Akshat Ramachandran, Souvik Kundu, and Tushar Krishna.
\newblock Polestar: Drift-aware cache calibration and token commitment for
  efficient inference of diffusion {LLMs}.
\newblock \emph{arXiv preprint arXiv:2607.14107}, 2026.

\bibitem[Lewkowycz et~al.(2022)Lewkowycz, Andreassen, Dohan, Dyer, Michalewski,
  Ramasesh, Slone, Anil, Schlag, Gutman-Solo, Wu, Neyshabur, Gur-Ari, and
  Misra]{lewkowycz2022minerva}
Aitor Lewkowycz, Anders Andreassen, David Dohan, Ethan Dyer, Henryk
  Michalewski, Vinay Ramasesh, Ambrose Slone, Cem Anil, Imanol Schlag, Theo
  Gutman-Solo, Yuhuai Wu, Behnam Neyshabur, Guy Gur-Ari, and Vedant Misra.
\newblock Solving quantitative reasoning problems with language models.
\newblock In \emph{Advances in Neural Information Processing Systems},
  volume~35, 2022.

\bibitem[Li et~al.(2023)Li, Liu, Lian, Yang, Dong, Kang, Zhang, and
  Keutzer]{li2023qdiffusion}
Xiuyu Li, Yijiang Liu, Long Lian, Huanrui Yang, Zhen Dong, Daniel Kang,
  Shanghang Zhang, and Kurt Keutzer.
\newblock {Q-Diffusion}: Quantizing diffusion models.
\newblock In \emph{Proceedings of the IEEE/CVF International Conference on
  Computer Vision (ICCV)}, pp.\  17535--17545, 2023.

\bibitem[Li et~al.(2021)Li, Gong, Tan, Yang, Hu, Zhang, Yu, Wang, and
  Gu]{li2021brecq}
Yuhang Li, Ruihao Gong, Xu~Tan, Yang Yang, Peng Hu, Qi~Zhang, Fengwei Yu, Wei
  Wang, and Shi Gu.
\newblock {BRECQ}: Pushing the limit of post-training quantization by block
  reconstruction.
\newblock In \emph{International Conference on Learning Representations}, 2021.

\bibitem[Lin et~al.(2025)Lin, Xu, Wu, Guo, Zhang, Lu, Wei, Zhang, and
  Sun]{lin2025quantmeetsdllm}
Haokun Lin, Haobo Xu, Yichen Wu, Ziyu Guo, Renrui Zhang, Zhichao Lu, Ying Wei,
  Qingfu Zhang, and Zhenan Sun.
\newblock Quantization meets d{LLMs}: A systematic study of post-training
  quantization for diffusion {LLMs}.
\newblock \emph{arXiv preprint arXiv:2508.14896}, 2025.

\bibitem[Lin et~al.(2024)Lin, Tang, Tang, Yang, Chen, Wang, Xiao, Dang, Gan,
  and Han]{lin2024awq}
Ji~Lin, Jiaming Tang, Haotian Tang, Shang Yang, Wei-Ming Chen, Wei-Chen Wang,
  Guangxuan Xiao, Xingyu Dang, Chuang Gan, and Song Han.
\newblock {AWQ}: Activation-aware weight quantization for on-device {LLM}
  compression and acceleration.
\newblock In \emph{Proceedings of Machine Learning and Systems}, 2024.

\bibitem[Liu et~al.(2025)Liu, Zhao, Fedorov, Soran, Choudhary, Krishnamoorthi,
  Chandra, Tian, and Blankevoort]{liu2025spinquant}
Zechun Liu, Changsheng Zhao, Igor Fedorov, Bilge Soran, Dhruv Choudhary,
  Raghuraman Krishnamoorthi, Vikas Chandra, Yuandong Tian, and Tijmen
  Blankevoort.
\newblock {SpinQuant}: {LLM} quantization with learned rotations.
\newblock In \emph{International Conference on Learning Representations}, 2025.

\bibitem[Liu et~al.(2026)Liu, Yang, Zhang, Chen, Zou, Wei, Wang, Zhu, and
  Zhang]{liu2025dllmcache}
Zhiyuan Liu, Yicun Yang, Yaojie Zhang, Junjie Chen, Chang Zou, Qingyan Wei,
  Shaobo Wang, Yichen Zhu, and Linfeng Zhang.
\newblock {dLLM-Cache}: Accelerating diffusion large language models with
  adaptive caching.
\newblock In \emph{International Conference on Machine Learning}, 2026.

\bibitem[Lou et~al.(2024)Lou, Meng, and Ermon]{lou2024sedd}
Aaron Lou, Chenlin Meng, and Stefano Ermon.
\newblock Discrete diffusion modeling by estimating the ratios of the data
  distribution.
\newblock In \emph{International Conference on Machine Learning}, 2024.

\bibitem[Ma et~al.(2025)Ma, Yu, Fang, and Wang]{ma2025dkvcache}
Xinyin Ma, Runpeng Yu, Gongfan Fang, and Xinchao Wang.
\newblock {dKV-Cache}: The cache for diffusion language models.
\newblock In \emph{Advances in Neural Information Processing Systems},
  volume~38, 2025.

\bibitem[Merity et~al.(2017)Merity, Xiong, Bradbury, and
  Socher]{merity2017wikitext}
Stephen Merity, Caiming Xiong, James Bradbury, and Richard Socher.
\newblock Pointer sentinel mixture models.
\newblock In \emph{International Conference on Learning Representations}, 2017.

\bibitem[Minderer et~al.(2021)Minderer, Djolonga, Romijnders, Hubis, Zhai,
  Houlsby, Tran, and Lucic]{minderer2021revisiting}
Matthias Minderer, Josip Djolonga, Rob Romijnders, Frances Hubis, Xiaohua Zhai,
  Neil Houlsby, Dustin Tran, and Mario Lucic.
\newblock Revisiting the calibration of modern neural networks.
\newblock In \emph{Advances in Neural Information Processing Systems},
  volume~34, 2021.

\bibitem[Nagel et~al.(2020)Nagel, Amjad, van Baalen, Louizos, and
  Blankevoort]{nagel2020adaround}
Markus Nagel, Rana~Ali Amjad, Mart van Baalen, Christos Louizos, and Tijmen
  Blankevoort.
\newblock Up or down? {A}daptive rounding for post-training quantization.
\newblock In \emph{International Conference on Machine Learning}, 2020.

\bibitem[Nie et~al.(2025)Nie, Zhu, You, Zhang, Ou, Hu, Zhou, Lin, Wen, and
  Li]{nie2025llada}
Shen Nie, Fengqi Zhu, Zebin You, Xiaolu Zhang, Jingyang Ou, Jun Hu, Jun Zhou,
  Yankai Lin, Ji-Rong Wen, and Chongxuan Li.
\newblock Large language diffusion models.
\newblock In \emph{Advances in Neural Information Processing Systems},
  volume~38, 2025.

\bibitem[Sahoo et~al.(2024)Sahoo, Arriola, Schiff, Gokaslan, Marroquin, Chiu,
  Rush, and Kuleshov]{sahoo2024mdlm}
Subham~Sekhar Sahoo, Marianne Arriola, Yair Schiff, Aaron Gokaslan, Edgar
  Marroquin, Justin~T. Chiu, Alexander Rush, and Volodymyr Kuleshov.
\newblock Simple and effective masked diffusion language models.
\newblock In \emph{Advances in Neural Information Processing Systems},
  volume~37, 2024.

\bibitem[Shang et~al.(2023)Shang, Yuan, Xie, Wu, and Yan]{shang2023ptq4dm}
Yuzhang Shang, Zhihang Yuan, Bin Xie, Bingzhe Wu, and Yan Yan.
\newblock Post-training quantization on diffusion models.
\newblock In \emph{Proceedings of the IEEE/CVF Conference on Computer Vision
  and Pattern Recognition (CVPR)}, 2023.

\bibitem[Shao et~al.(2024)Shao, Chen, Zhang, Xu, Zhao, Li, Zhang, Gao, Qiao,
  and Luo]{shao2024omniquant}
Wenqi Shao, Mengzhao Chen, Zhaoyang Zhang, Peng Xu, Lirui Zhao, Zhiqian Li,
  Kaipeng Zhang, Peng Gao, Yu~Qiao, and Ping Luo.
\newblock {OmniQuant}: Omnidirectionally calibrated quantization for large
  language models.
\newblock In \emph{International Conference on Learning Representations}, 2024.

\bibitem[Shi et~al.(2024)Shi, Han, Wang, Doucet, and Titsias]{shi2024md4}
Jiaxin Shi, Kehang Han, Zhe Wang, Arnaud Doucet, and Michalis~K. Titsias.
\newblock Simplified and generalized masked diffusion for discrete data.
\newblock In \emph{Advances in Neural Information Processing Systems},
  volume~37, 2024.

\bibitem[Sun et~al.(2025)Sun, Liu, Bai, Bao, Zhao, Li, Hu, Yu, Hou, Yuan,
  Jiang, Liu, and Yao]{sun2025flatquant}
Yuxuan Sun, Ruikang Liu, Haoli Bai, Han Bao, Kang Zhao, Yuening Li, Jiaxin Hu,
  Xianzhi Yu, Lu~Hou, Chun Yuan, Xin Jiang, Wulong Liu, and Jun Yao.
\newblock {FlatQuant}: Flatness matters for {LLM} quantization.
\newblock In \emph{International Conference on Machine Learning}, 2025.

\bibitem[Vaswani et~al.(2017)Vaswani, Shazeer, Parmar, Uszkoreit, Jones, Gomez,
  Kaiser, and Polosukhin]{vaswani2017attention}
Ashish Vaswani, Noam Shazeer, Niki Parmar, Jakob Uszkoreit, Llion Jones,
  Aidan~N. Gomez, {\L}ukasz Kaiser, and Illia Polosukhin.
\newblock Attention is all you need.
\newblock In \emph{Advances in Neural Information Processing Systems},
  volume~30, 2017.

\bibitem[Wu et~al.(2026{\natexlab{a}})Wu, Zhang, Xue, Diao, Fu, Liu, Molchanov,
  Luo, Han, and Xie]{wu2026fastdllmv2}
Chengyue Wu, Hao Zhang, Shuchen Xue, Shizhe Diao, Yonggan Fu, Zhijian Liu,
  Pavlo Molchanov, Ping Luo, Song Han, and Enze Xie.
\newblock Fast-d{LLM} v2: Efficient block-diffusion {LLM}.
\newblock In \emph{International Conference on Learning Representations},
  2026{\natexlab{a}}.

\bibitem[Wu et~al.(2026{\natexlab{b}})Wu, Zhang, Xue, Liu, Diao, Zhu, Luo, Han,
  and Xie]{wu2025fastdllm}
Chengyue Wu, Hao Zhang, Shuchen Xue, Zhijian Liu, Shizhe Diao, Ligeng Zhu, Ping
  Luo, Song Han, and Enze Xie.
\newblock Fast-d{LLM}: Training-free acceleration of diffusion {LLM} by
  enabling {KV} cache and parallel decoding.
\newblock In \emph{International Conference on Learning Representations},
  2026{\natexlab{b}}.

\bibitem[Xiao et~al.(2023)Xiao, Lin, Seznec, Wu, Demouth, and
  Han]{xiao2023smoothquant}
Guangxuan Xiao, Ji~Lin, Mickael Seznec, Hao Wu, Julien Demouth, and Song Han.
\newblock {SmoothQuant}: Accurate and efficient post-training quantization for
  large language models.
\newblock In \emph{International Conference on Machine Learning}, 2023.

\bibitem[Xu et~al.(2026)Xu, Chen, and Yang]{xu2025dllmquant}
Chen Xu, Zhixuan Chen, and Dawei Yang.
\newblock {DLLMQuant}: A post-training quantization framework tailored for
  diffusion-based large language models.
\newblock In \emph{International Conference on Machine Learning}, 2026.

\bibitem[Yan et~al.(2026)Yan, Wang, Wan, Yu, and Tsang]{yan2026starquant}
Xin Yan, Aqiang Wang, Zhenglin Wan, Xingrui Yu, and Ivor Tsang.
\newblock {STaR-Quant}: State-time consistent post-training quantization for
  diffusion large language models.
\newblock \emph{arXiv preprint arXiv:2606.04945}, 2026.

\bibitem[Yang et~al.(2024)Yang, Yang, Zhang, Hui, et~al.]{yang2024qwen25}
An~Yang, Baosong Yang, Beichen Zhang, Binyuan Hui, et~al.
\newblock Qwen2.5 technical report.
\newblock \emph{arXiv preprint arXiv:2412.15115}, 2024.

\bibitem[Ye et~al.(2025)Ye, Xie, Zheng, Gao, Wu, Jiang, Li, and
  Kong]{ye2025dream}
Jiacheng Ye, Zhihui Xie, Lin Zheng, Jiahui Gao, Zirui Wu, Xin Jiang, Zhenguo
  Li, and Lingpeng Kong.
\newblock Dream 7{B}: Diffusion large language models.
\newblock \emph{arXiv preprint arXiv:2508.15487}, 2025.

\bibitem[Zhang et~al.(2026)Zhang, Li, Yan, Qin, Guo, and
  Zhang]{zhang2026quantdllm}
Tianao Zhang, Zhiteng Li, Xianglong Yan, Haotong Qin, Yong Guo, and Yulun
  Zhang.
\newblock Quant-d{LLM}: Post-training extreme low-bit quantization for
  diffusion large language models.
\newblock In \emph{International Conference on Learning Representations}, 2026.

\end{thebibliography}
\bibliographystyle{iclr2027_conference}

\clearpage
\appendix
\counterwithin{table}{section}
\counterwithin{figure}{section}

\section{On-Policy Paired Measurement of the Parallelism Tax}
\label{app:paired}

Comparing the forward passes of two free-running decoders mixes two effects: quantization lets fewer positions clear $\tau$ at each step, and the two decoders visit different states after their first disagreement. Only the first is caused by quantization error. After reporting the end-to-end tax of every configuration, this appendix isolates the first effect on LLaDA-8B-Instruct with QuaRot, the configuration used throughout Appendices~A--D.

\begin{figure}[ht]
  \centering
  \includegraphics[width=0.68\linewidth]{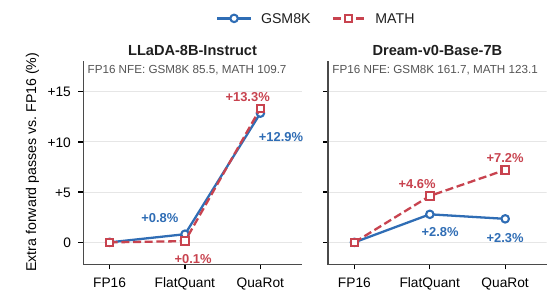}
  \caption{End-to-end parallelism tax on both tasks: extra forward passes per response relative to FP16 at $T=1$ (Fast-dLLM, dual cache, $\tau=0.9$). Its magnitude depends on the model, the quantizer, and the task.\looseness=-1}
  \label{fig:tax-tasks}
\end{figure}

\paragraph{End-to-end tax.}
\Cref{fig:tax-tasks} extends \cref{fig:obs}c to MATH. The tax appears on every model, quantizer, and task, with a magnitude that ranges from 0.1\% (FlatQuant on LLaDA, MATH) to 13.3\% (QuaRot on LLaDA, MATH).

\paragraph{Setup.}
The quantized model has exclusive control of decoding and produces states $x_1,\dots,x_S$. At every step, a full-precision shadow runs one read-only forward pass on bit-identical input; for QuaRot, it is a copy of the model taken after the rotations are folded in and before the weights are quantized. It keeps its own KV cache and never commits, so any difference between the two arms at a state is due to numerical precision alone.

\paragraph{Estimator.}
The width of model $m$ at state $x$ is the number of tokens it would write,
\begin{equation*}
w^m(x)=\max\big(\,|\{i\in\mathcal M(x):p^m_i(x)\ge\tau\}|,\ 1\big),
\end{equation*}
where $\mathcal M(x)$ is the set of masked positions of the active block and $\max(\cdot,1)$ encodes the forced commitment. Width is a count rather than a set because the forced rule fixes how many positions are written, not which: each arm forces its own most confident position, and a mismatch in position costs no parallelism. Averaging over the trajectory, $W^m=\frac1S\sum_{t}w^m(x_t)$, the paired tax is $\hat{\mathcal T}=W^{\mathrm{fp}}/W^{\mathrm q}-1$: generating $L$ tokens takes $S=L/W$ steps, so $\hat{\mathcal T}$ is the relative excess of steps at the quantized width. It uses only same-state comparisons, and the identity $\sum_t w^{\mathrm q}(x_t)=L$ checks the recording exactly.\looseness=-1

\begin{table}[ht]
\centering\small
\caption{The paired estimate and the end-to-end tax, measured on disjoint data.}
\label{tab:paired}
\begin{tabular}{lll}
\toprule
 & Paired on-policy & Independent end-to-end \\
\midrule
Data & 32 prompts, 2,910 steps, 47,919 paired positions & 400 questions, two complete runs \\
Reading & $W^{\mathrm{fp}}=3.193$, $W^{\mathrm q}=2.815$ & $\mathrm{NFE}^{\mathrm{fp}}=85.23$, $\mathrm{NFE}^{\mathrm q}=97.64$ \\
Tax & $\hat{\mathcal T}=13.4\%$ & $\mathrm{NFE}^{\mathrm q}/\mathrm{NFE}^{\mathrm{fp}}-1=14.6\%$ \\
\bottomrule
\end{tabular}
\end{table}

\paragraph{The tax is per-step width loss.}
The two estimates in \cref{tab:paired} share no data and differ by 1.2 points. Since $\hat{\mathcal T}$ consists of per-step width loss only, the forward-pass gap is explained by fewer commitments per step rather than by trajectory divergence. The small residual is expected: $\hat{\mathcal T}$ is conditioned on the states the quantized policy visits, whereas the full-precision decoder follows its own trajectory, on which positions deferred at one step can be released together later.

\paragraph{Withheld positions keep their content.}
Let $R$ be the positions the shadow would commit but the quantized model withholds ($N=1{,}700$). The two models predict the same token on 97.94\% of $R$ (96.84\% at build steps, 98.16\% at reuse steps), against 68.31\% over all masked positions: quantization perturbs predictions broadly, but mostly at low confidence, where no commitment is decided. Because $R$ is selected by ``full precision confident, quantized not'', we also condition on the quantized side only, using the decision band $p^{\mathrm q}\in[\tau-0.15,\tau+0.05]$; agreement is 97.52\%. The withheld parallelism thus consists of commitments whose content is not in dispute, which is what makes it recoverable. Whether recovering it is accuracy-neutral is a separate question, answered against the quantized model at $T=1$ (\cref{fig:temp}).

\section{Where Quantization Compresses the Logits}
\label{app:logits}

\paragraph{Probe.}
At the first build step of block~0, all generation positions are masked and both models receive bit-identical inputs, so their logits can be paired position by position (GSM8K, 400 questions; HumanEval, 164). Two independent estimators quantify the compression. The geometric one sorts each position's logits and fits a single through-origin scale $d^{\mathrm Q}_j=\kappa_{\mathrm{gap}}\,d^{\mathrm{FP}}_j$ to the gaps $d_j=L_{(1)}-L_{(j)}$, $j=2,\dots,64$; it sees only shape. The behavioural one solves exactly, over the full vocabulary, for the temperature $\theta$ with $\max\mathrm{softmax}(L^{\mathrm{FP}}/\theta)=p^{\mathrm Q}_1$; it sees only the outcome, and $1/\theta$ is the equivalent compression.

\begin{table}[ht]
\centering\small
\caption{Compression by full-precision confidence (medians per stratum). QR: QuaRot; NR: W4A4 without rotation (group size 128); G: GSM8K; H: HumanEval.}
\label{tab:kappa}
\begin{tabular}{cc ccc ccc c}
\toprule
 & & \multicolumn{3}{c}{positions} & \multicolumn{3}{c}{$\kappa_{\mathrm{gap}}$} & $1/\theta$ \\
\cmidrule(lr){3-5}\cmidrule(lr){6-8}\cmidrule(lr){9-9}
$p^{\mathrm{FP}}_1$ & $p^{\mathrm{FP}}_1\!\to p^{\mathrm Q}_1$ (QR-G) & QR-G & NR-G & QR-H & QR-G & NR-G & QR-H & QR-G \\
\midrule
$[0.0,0.3)$ & $0.115\to0.110$ & 9,306 & 9,299 & 4,511 & 0.974 & 0.924 & 1.002 & 0.989 \\
$[0.3,0.6)$ & $0.419\to0.342$ & 2,533 & 2,533 & 648 & 0.908 & 0.883 & 0.906 & 0.881 \\
$[0.6,0.8)$ & $0.678\to0.559$ & 740 & 748 & 86 & 0.905 & 0.890 & 0.885 & 0.857 \\
$[0.8,0.9)$ & $0.847\to0.729$ & 153 & 151 & 0 & 0.895 & 0.899 & -- & 0.842 \\
$[0.9,1.0]$ & $0.936\to0.872$ & 68 & 69 & 0 & 0.899 & 0.895 & -- & 0.835 \\
\bottomrule
\end{tabular}
\end{table}

\begin{table}[ht]
\centering\small
\caption{Compression by scale (GSM8K). Rank-wise rows (top-2 to top-64) pool positions with $p^{\mathrm{FP}}_1\ge0.3$ ($n=3{,}494$); the remaining rows use all positions ($n=12{,}800$). $\kappa_{\mathrm{all}}$ is the OLS slope of $L^{\mathrm Q}_u=\alpha+\kappa_{\mathrm{all}}L^{\mathrm{FP}}_u+\nu_u$ over 1,024 token ids fixed by a seed and independent of both models; $\sigma_\nu/s_{\mathrm{FP}}=\sqrt{(s_{\mathrm Q}/s_{\mathrm{FP}})^2-\kappa_{\mathrm{all}}^2}$ follows from $s^2_{\mathrm Q}=\kappa_{\mathrm{all}}^2s^2_{\mathrm{FP}}+\sigma^2_\nu$.}
\label{tab:spectrum}
\begin{tabular}{llcc}
\toprule
Scale & Quantity & QuaRot & NR (g128) \\
\midrule
top-2 & $d^{\mathrm Q}_2/d^{\mathrm{FP}}_2$ & 0.740 & 0.743 \\
top-3 & $d^{\mathrm Q}_3/d^{\mathrm{FP}}_3$ & 0.802 & 0.803 \\
top-5 & $d^{\mathrm Q}_5/d^{\mathrm{FP}}_5$ & 0.843 & 0.838 \\
top-10 & $d^{\mathrm Q}_{10}/d^{\mathrm{FP}}_{10}$ & 0.878 & 0.863 \\
top-64 & $\kappa_{\mathrm{gap}}$ & 0.906 & 0.885 \\
1,024 fixed tokens & $\kappa_{\mathrm{all}}$ & 0.965 & 0.942 \\
full vocabulary & $s_{\mathrm Q}/s_{\mathrm{FP}}$ & 1.007 & 0.992 \\
full vocabulary & $\sigma_\nu/s_{\mathrm{FP}}$ & 0.288 & 0.311 \\
\bottomrule
\end{tabular}
\end{table}

\paragraph{Compression is systematic.}
Confidence drops in every stratum of \cref{tab:kappa}, and in the decision-relevant strata ($p^{\mathrm{FP}}_1\ge0.3$) it has the same size for both quantizers and both tasks, so it is a property of the model and quantizer rather than of one rotation scheme or prompt distribution. Its consequence shows on the probe itself: of the positions full precision would commit, only 23 of 68 (QuaRot) and 46 of 69 (no rotation) still clear $\tau$ after quantization, a small-sample illustration of direction rather than a measure of the tax.

\paragraph{Compression is concentrated at the peak.}
\Cref{tab:spectrum} shows compression weakening monotonically as the scale widens: the gap between the peak and its strongest competitor shrinks to 0.74, while the slope over model-independent tokens is close to one and the standard deviation of the whole logit vector is unchanged. Quantization adds nearly independent noise that offsets the shrinkage in the bulk of the vocabulary, but thins exactly the few gaps on which the maximum softmax probability depends.\looseness=-1

\paragraph{Implication for the temperature.}
Dividing the logits by $T$ scales every gap by $1/T$, so a temperature can undo this compression. Because the compression depends on scale and on the confidence stratum, no single number in these tables is the right temperature, and we calibrate it on the commit rate instead (\cref{sec:temp}).

\section{Temperature Ladder: Mechanism and Premature Commitment}
\label{app:ladder}

\begin{table}[ht]
\centering\small
\caption{Instrumented temperature ladder (GSM8K test set; temperature on reuse steps). NFE and single-sequence throughput are relative to $T=1$ of the same run. $F$: forced steps per question; $\bar k$: mean width of threshold steps. Early release: share of generated positions committed earlier than at $T=1$. First-release agreement: share of questions whose token released at the first early release equals the one the $T=1$ schedule eventually writes. Slot agreement: share of output slots identical to $T=1$ (trailing EOS padding excluded).}
\label{tab:ladder}
\setlength{\tabcolsep}{4.5pt}
\begin{tabular}{cccccccc}
\toprule
$T$ & NFE & $F$ & $\bar k$ & throughput & early release & first-release agr. & slot agr. \\
\midrule
1    & --          & 45.81 & 4.02 & --          & --     & --     & --     \\
0.94 & $-5.9\%$  & 40.74 & 4.18 & $+6.0\%$  & 5.0\%  & 99.4\% & 70.4\% \\
0.87 & $-12.6\%$ & 35.06 & 4.36 & $+12.9\%$ & 11.2\% & 99.0\% & 52.1\% \\
0.84 & $-15.3\%$ & 32.69 & 4.44 & $+14.1\%$ & 14.0\% & 98.7\% & 45.7\% \\
\bottomrule
\end{tabular}
\end{table}

\paragraph{Recovery reverses the cause.}
With $K$ generated positions, $F$ forced steps, and threshold steps of mean width $\bar k$, the number of forward passes is $\mathrm{NFE}=K/\bar k+F\,(1-1/\bar k)$. The tax raises $F$ and lowers $\bar k$; the temperature does the opposite. From $T=1$ to $T=0.84$, forced steps fall by 28.6\% and the threshold width rises by 10.4\%, i.e., the merges that quantization suppressed happen again. Forward passes and measured throughput move together at every temperature (e.g., $-12.6\%$ and $+12.9\%$ at $T=0.87$), so in the single-sequence setting NFE is a faithful proxy for latency.

\paragraph{Premature commitment is rare and mostly one step early.}
Since \cref{prop:temp} fixes what is written at a given state, the only risk the temperature introduces is committing a position before its context is complete. We measure it along the $T=1$ trajectory: at every visited state and for every position that $T=0.87$ would release there, we compare the token released at that state with the token the $T=1$ schedule eventually writes. Both are recorded values, so the comparison involves no approximation; it characterizes the intervention itself rather than its downstream course, since the two schedules separate after the first early release. Four in five such releases are advanced by a single step, and these agree in 99.9\% of cases. Agreement decreases monotonically with the advance, to about one half, but advances of that depth account for less than 0.1\% of releases. The first-release agreement in \cref{tab:ladder} is restricted to the step at which both schedules provably occupy the same state and is therefore exact and on-policy.

\paragraph{Why outputs still change.}
Slot agreement falls from 70.4\% to 45.7\% without contradicting \cref{prop:temp}: the proposition concerns what is written at a given state, whereas slot agreement compares the endpoints of two different trajectories. Once a position is committed earlier, its neighbours face a different context at the same step, so outputs diverge while every write remains the argmax of its own state. \Cref{prop:temp} rules out that the temperature makes the model write wrongly, not that it changes the output.

\section{Compression and Suppression Along the Trajectory}
\label{app:strata}

\begin{table}[ht]
\centering\small
\caption{On-policy paired recording (32 GSM8K prompts, 2,910 steps, 47,919 paired positions), stratified by the filled fraction $\rho$ of the active block ($\rho=0$ at build steps). FP / Q commit: positions above $\tau$ in each arm. Suppression: share of full-precision commitments the quantized arm withholds. Share of pool: share of all 1,700 withheld positions. $\kappa^\star$: median of the per-position equivalent compression, solved exactly.}
\label{tab:strata}
\setlength{\tabcolsep}{4.5pt}
\begin{tabular}{lcccccc}
\toprule
Stratum & paired positions & FP commit & Q commit & suppression & share of pool & $\kappa^\star$ \\
\midrule
build ($\rho=0$)          & 8,192  & 2,362 & 2,157 & 12.1\% & 16.8\% & 0.930 \\
reuse, $\rho\in[0,.25)$   & 15,293 & 1,571 & 1,274 & 24.8\% & 22.9\% & 0.966 \\
reuse, $\rho\in[.25,.50)$ & 12,906 & 1,607 & 1,324 & 24.1\% & 22.8\% & 0.930 \\
reuse, $\rho\in[.50,.75)$ & 7,991  & 1,548 & 1,231 & 26.5\% & 24.1\% & 0.921 \\
reuse, $\rho\in[.75,1)$   & 3,537  & 1,033 & 873   & 22.0\% & 13.4\% & 0.912 \\
\midrule
all                       & 47,919 & 8,121 & 6,859 & 20.9\% & 100\%  & 0.939 \\
\bottomrule
\end{tabular}
\end{table}

\paragraph{The block-0 probe is representative.}
\Cref{app:logits} reads the compression at the first build step of block~0, the only point at which both arms see bit-identical inputs, whereas the temperature acts on reuse steps. Along a whole trajectory, the median $\kappa^\star$ spans only 0.912--0.966 across strata (\cref{tab:strata}), and the build-step value (0.930) lies in the same neighbourhood as the trajectory median (0.939) and the probe's decision-relevant values (0.895--0.908).

\paragraph{Reading the table.}
The suppression ratios are position-level upper bounds, since at a forced step each arm commits its own most confident position without any loss of parallelism; the tax itself is measured by the counting estimator of \cref{app:paired}.

\section{Design Choices of the Commit Temperature}
\label{app:design}

\paragraph{Build steps: least to gain, most to lose.}
\Cref{tab:build} combines the paired recording of \cref{app:paired,app:strata}. A response has one build step per block, 8 of roughly 90 forward passes, and these steps hold only 285 of the 1,700 withheld commitments (16.8\%), because build steps are the least suppressed stratum. Releasing a withheld commitment writes a token that full precision would not write with probability one minus the agreement rate, 3.2\% at build steps against 1.8\% at reuse steps; weighted by the number of releases, build steps would contribute 25.7\% of the expected miswrites. Each of them, moreover, is written into a fully masked block and becomes context for the other 31 positions in every later step, whereas a release at a late reuse step affects only the few positions still masked. Build steps therefore keep $T=1$.

\begin{table}[ht]
\centering\small
\caption{Expected benefit and cost of releasing the withheld commitments, by step type (paired recording of \cref{app:paired}). Expected miswrites: withheld positions $\times$ (1 $-$ agreement).}
\label{tab:build}
\begin{tabular}{lccccc}
\toprule
Step type & withheld positions & share of gain & agreement & expected miswrites & share of miswrites \\
\midrule
build & 285   & 16.8\% & 96.84\% & 9.0  & 25.7\% \\
reuse & 1,415 & 83.2\% & 98.16\% & 26.0 & 74.3\% \\
\bottomrule
\end{tabular}
\end{table}

\paragraph{Relation to lowering $\tau$.}
With two candidates, $p^{T}\ge\tau$ holds exactly when $\Delta\ge T m_\tau$, so on the original confidence a global temperature acts like the threshold $\tau_{\mathrm{eff}}=\sigma(Tm_\tau)$, which is 0.858 at $T=0.82$ and $\tau=0.9$. To first order, the temperature is therefore a reparametrization of $\tau$, and \cref{tab:tau} confirms that the two cannot be told apart at matched forward passes. Interpolating the temperature ladder to the forward-pass count of $\tau=0.82$ predicts its forced steps and threshold width to within 0.1\% (30.84 against 30.82, and 4.566 against 4.564), and both settings reach the same accuracy. We therefore do not claim a new decoding knob. What~(\ref{eq:tmatch}) contributes is how to set this knob without labels: the value it returns reads as the quantizer's compression ($T=\kappa$ under uniform contraction, \cref{prop:temp}), is calibrated from forward-pass counts alone and carries over across tasks (\cref{sec:exp-temp}), and leaves $\tau$ at the value its deployer chose for the full-precision model.

\begin{table}[ht]
\centering\small
\caption{Lowering the temperature and lowering the threshold (LLaDA-8B-Instruct with QuaRot, GSM8K, 400 questions; same instrumentation as \cref{tab:paired}; the temperature acts on reuse steps, the threshold on every step). $F$: forced steps per question; $\bar k$: mean width of threshold steps.}
\label{tab:tau}
\begin{tabular}{lcccc}
\toprule
Setting & NFE & $F$ & $\bar k$ & accuracy (\%) \\
\midrule
FP16, $\tau=0.9$              & 85.23 & 39.05 & 4.698 & 78.25 \\
W4A4, $\tau=0.9$, $T=1$       & 97.64 & 46.16 & 4.076 & 76.50 \\
W4A4, $\tau=0.9$, $T=0.87$    & 85.04 & 35.13 & 4.426 & 76.00 \\
W4A4, $\tau=0.9$, $T=0.82$    & 80.88 & 31.48 & 4.545 & 75.75 \\
W4A4, $\tau=0.82$, $T=1$      & 80.16 & 30.82 & 4.564 & 75.75 \\
\bottomrule
\end{tabular}
\end{table}

\paragraph{Global versus stratum-wise temperatures.}
For every position there is an exact restoring temperature $\kappa^\star$ (\cref{tab:strata}), but computing it requires the full-precision logits, which are unavailable once the model is deployed in quantized form. The realizable alternative assigns each stratum of \cref{tab:strata} its own temperature, its median $\kappa^\star$. We compare the two on the paired recording at matched budgets of premature commitments, i.e., positions the corrected quantized arm would commit but full precision would not, and count the withheld commitments each recovers. At three budgets, the stratum-wise temperatures recover 0.99, 1.01, and 1.00 times as many as the global one. The stratum carries information about the compression, but not about which positions lie just below the threshold, which is all a threshold decoder acts on.

\section{Proof of Proposition~\ref{prop:temp} and the Premise of the Reach}
\label{app:proof}

\paragraph{Proof of \cref{prop:temp}.}

(i) The argmax is invariant to positive scaling, so dividing the logits by $T>0$ leaves every written token unchanged. (ii) Replacing $\Delta_v$ by $\Delta_v/T$ in~(\ref{eq:margin}) shows that $p^{T}$ grows as $T$ decreases, so the threshold set only grows, and a non-empty one commits at least as many positions as a forced commitment; with two candidates, $p<\tau\le p^{T}$ is equivalent to $Tm_\tau\le\Delta<m_\tau$. (iii) The softmax is shift-invariant, so $z^{\mathrm q}=\kappa z+b\mathbf 1$ gives $\mathrm{softmax}(z^{\mathrm q}/\kappa)=\mathrm{softmax}(z+b\mathbf 1/\kappa)=\mathrm{softmax}(z)$.

\paragraph{Premise of the reach.}
The square root in~(\ref{eq:reach}) is the root-mean-square accumulation of approximately independent perturbations across the passes that read a committed token. We verified the independence premise: without the shared per-layer bias, cross-position error terms are 3.6\% of the diagonal ones. The choice of the concave map matters little: $\log(1{+}\cdot)$ in place of the square root gives a prior with correlation 0.9975.\looseness=-1

\section{Forced Commitments and Wrong Answers}
\label{app:forced}

We test whether the errors that enter through forced commitments decide answers. All numbers come from the same-state recordings of \cref{fig:obs}b (W4A4 FlatQuant, first 400 GSM8K test questions, Fast-dLLM with dual cache, $\tau=0.9$), except the example at the end, which uses W4A4 without rotation (group size 128, \cref{app:logits}). A question is \emph{broken} if full precision answers it correctly and W4A4 does not, and \emph{kept} if both do; a \emph{forced flip} is a token flip at a forced commitment.

\paragraph{Wrong answers contain more forced flips.}
\Cref{tab:forced-count} counts forced flips per question. Many single forced flips are harmless; what separates broken from kept questions is how many they contain.

\begin{table}[H]
\centering\small
\caption{Average number of forced flips per question (number of questions in parentheses). CI: bootstrap over questions.}
\label{tab:forced-count}
\setlength{\tabcolsep}{4pt}
\begin{tabular}{lccccc}
\toprule
 & W4A4 wrong & W4A4 right & Broken & Kept & Broken $-$ Kept (95\% CI) \\
\midrule
LLaDA-8B & 3.08 (88) & 1.99 (312) & 3.36 (33) & 1.95 (282) & $+1.4$ [0.6, 2.4] \\
Dream-7B & 3.90 (121) & 2.81 (279) & 4.36 (36) & 2.70 (252) & $+1.7$ [1.0, 2.4] \\
\bottomrule
\end{tabular}
\end{table}

\paragraph{The risk of a wrong answer grows with the number of forced flips.}
Among questions that full precision answers correctly, the share that W4A4 gets wrong rises from 0--4\% without forced flips to 28--29\% with five or more (\cref{tab:forced-risk}). The effect holds after controlling for question length and the number of forced commitments: each additional forced flip multiplies the odds of a wrong answer by 1.25 on LLaDA and 1.24 on Dream (95\% CI 1.01--1.57 and 1.01--1.55).

\begin{table}[H]
\centering\small
\caption{Share of questions that full precision answers correctly but W4A4 gets wrong, by number of forced flips.}
\label{tab:forced-risk}
\begin{tabular}{lccccc}
\toprule
Forced flips & 0 & 1 & 2 & 3--4 & $\ge5$ \\
\midrule
LLaDA-8B & 4\% (2/47) & 8\% (7/84) & 5\% (4/86) & 18\% (13/73) & 28\% (7/25) \\
Dream-7B & 0\% (0/23) & 5\% (3/56) & 7\% (4/59) & 14\% (13/95) & 29\% (16/55) \\
\bottomrule
\end{tabular}
\end{table}

\paragraph{The first surviving error enters through a forced commitment.}
In broken answers, the first token flip that survives in the output comes from a commitment that bypassed the threshold in 29 of 31 cases on LLaDA and in all 36 on Dream (32 forced commitments and 4 block-initial commitments, which Dream's decoder makes once per block without a threshold test). Counting placement flips as well, forced commitments carry 99.5\% (LLaDA) and 96.7\% (Dream) of the permanent flips.

\paragraph{Correcting only forced commitments restores the full-precision answers.}
We reran W4A4 on the same questions with one change: at every forced commitment (on Dream, also at block-initial commitments), it used the full-precision choice of position and token for the same state; all other commitments stayed quantized, and the decoding schedule was unchanged. This repairs about nine in ten broken answers (\cref{tab:forced-fix}), so forced commitments are the main channel through which quantization changes answers. As a deployed fix, however, recomputing forced steps would require a second pass for each of them, since a step is known to be forced only after its W4A4 pass; forced steps make up 44\% of all forward passes on LLaDA and 68\% on Dream (including block-initial commitments), far more than the tax of \cref{sec:obs2}. \method{} acts on the same decisions at calibration time, at no inference cost.

\begin{table}[H]
\centering\small
\caption{Rerunning W4A4 with only its forced commitments (on Dream, also its block-initial commitments) replaced by the full-precision choice (400 GSM8K questions).}
\label{tab:forced-fix}
\begin{tabular}{lcc}
\toprule
 & LLaDA-8B & Dream-7B \\
\midrule
Questions broken by quantization & 33 & 36 \\
\quad of which answered correctly again & 30 (91\%) & 32 (89\%) \\
Questions whose outcome differs from full precision (before $\to$ after) & 63 $\to$ 12 & 63 $\to$ 15 \\
Agreement with full precision (before $\to$ after) & 84\% $\to$ 97\% & 84\% $\to$ 96\% \\
\bottomrule
\end{tabular}
\end{table}

\paragraph{Why a near tie can still break an answer.}
At a near tie, either token is often fine locally. A tie becomes harmful when it decides how the solution continues: a diffusion model can write any position of the active block, so one forced commitment can fix the form of the next lines. GSM8K test question~197 shows this. Both runs first write the identical line \mbox{``Number of children = 23786 - 8417 - 9092''} and its line break, which puts them in the same decoding state. At the next forward pass, no position clears the threshold, so each run writes only its most confident position (highlighted; \texttt{[MASK]}: a position not yet written; $\hookleftarrow$: line break):

\medskip
\begingroup\footnotesize\setlength{\tabcolsep}{2.5pt}\renewcommand{\arraystretch}{1.3}
\newcommand{\msk}{\texttt{[MASK]}}
\newcommand{\hlt}[1]{\fcolorbox{blue}{yellow!45}{#1}}
\noindent\begin{tabular}{l l cccccc l}
 & & \multicolumn{6}{c}{position in the generation window} & \\
 & & 97 & 98 & 99 & 100 & 101 & 102 & \\
\midrule
Shared state before the pass & \texttt{\dots\ 9092} & $\hookleftarrow$ & \msk & \msk & \msk & \msk & \msk & \texttt{\dots} \\
Full precision after the pass & \texttt{\dots\ 9092} & $\hookleftarrow$ & \hlt{$\hookleftarrow$} & \msk & \msk & \msk & \msk & \texttt{\dots} \\
W4A4 after the pass & \texttt{\dots\ 9092} & $\hookleftarrow$ & \msk & \msk & \msk & \hlt{\texttt{=}} & \msk & \texttt{\dots} \\
\end{tabular}
\endgroup
\medskip

\noindent\begin{minipage}{\linewidth}
The two outputs then read:\\[6pt]
\textbf{Full precision} (correct)\\[1pt]
{\footnotesize\ttfamily
\hspace*{1.5em}Number of children = 23786 - 8417 - 9092\\
\hspace*{1.5em}\fcolorbox{blue}{yellow!45}{$\hookleftarrow$}\\
\hspace*{1.5em}Now, let's do the subtraction:\\[3pt]
\hspace*{1.5em}23786 - 8417 = 15369\\
\hspace*{1.5em}15369 - 9092 = 6277}\\[5pt]
\textbf{W4A4} (wrong)\\[1pt]
{\footnotesize\ttfamily
\hspace*{1.5em}Number of children = 23786 - 8417 - 9092\\
\hspace*{1.5em}Number of children \fcolorbox{blue}{yellow!45}{=} 23786 - 16419\\
\hspace*{1.5em}Number of children = 7277}
\end{minipage}
\medskip

Full precision writes a second line break at position 98 (confidence 0.77, against 0.22 for ``Number''), leaving a blank line, and then works step by step. W4A4 faces a near tie between two positions, ``Number'' at position 98 (0.62) and ``='' three positions ahead, at position 101 (0.66), and writes the ``='' first. Once ``='' is fixed, the next pass writes ``Number of children'' in front of it, each token at a confidence of at least 0.999, so the next line is committed to a single equation. The sum $8417+9092=17509$ is never written out and comes out as 16419, and the answer is 7277 instead of 6277. Both continuations are coherent, but the W4A4 one moves an addition from the page into the model's internal computation, which is exactly what quantization degrades. Near-tied choices are therefore not equally safe for a quantized model, which is why the risk grows with the number of forced flips and why the prior weights each position by the reach of its commitment.
\section{How \method{} and the Commit Temperature Divide the Work}
\label{app:division}

Quantization perturbs the logits, and the threshold sorts the affected decisions by margin. Fragile decisions with small margins are forced even at full precision and can flip, which costs accuracy (Observation~1); \method{} calibrates the model to perturb them less. Robust decisions just below the threshold are only delayed, which costs forward passes (Observation~2); the temperature restores the confidence scale so that they pass on time. To check this division, we decoded the 400 GSM8K questions of \cref{fig:obs}b with W4A4 QuaRot at $T=1$ and checked, at every forced step, whether $T_{\mathrm{match}}=0.873$ would let a position clear the threshold; a full-precision copy run on the same states marks token flips.\looseness=-1

\begin{table}[H]
\centering\small
\caption{Commitments in the returned output by type (W4A4 QuaRot at $T=1$, 400 GSM8K questions). Flip rate: share of commitments whose token differs from full precision.}
\label{tab:division}
\begin{tabular}{lcccc}
\toprule
Commitment & \makecell{Share of\\commitments} & \makecell{Share of\\token flips} & Flip rate & \makecell{FP16 confidence\\(median)} \\
\midrule
Above the threshold & 79.8\% & 8.3\% & 0.29\% & 0.994 \\
Forced, released by the temperature & 3.3\% & 2.2\% & 1.9\% & 0.876 \\
Forced, still forced & 16.9\% & 89.4\% & 14.6\% & 0.689 \\
\bottomrule
\end{tabular}
\end{table}

The temperature releases only forced commitments just below the threshold, which rarely flip (\cref{tab:division}). The forced commitments that carry the errors lie far below the threshold, stay forced, and account for 97.6\% of the flips at forced commitments; they are exactly where the prior of \method{} puts its weight, since it grows as the full-precision margin shrinks. This is why the temperature removes forced steps (\cref{tab:ladder}) while leaving accuracy essentially unchanged (\cref{fig:temp}), and why it cannot replace calibration: it changes when a position is committed, never what is written (\cref{prop:temp}).

\section{Checks of the Flip-Probability Model}
\label{app:flipmodel}

Both the risk model~(\ref{eq:flip}) and the reach~(\ref{eq:reach}) serve only to rank positions, since~(\ref{eq:prior}) keeps nothing but their order. We checked each part of~(\ref{eq:flip}) against the flips recorded for \cref{fig:obs}b (W4A4 FlatQuant, first 400 GSM8K test questions; 102,400 commitments on LLaDA and 99,200 on Dream).

\paragraph{Smaller margins flip more often.}
Ranked by the full-precision margin from which $\pi$ is computed, a commitment whose token flips is rated riskier than one whose token does not in 98.7\% of such pairs on LLaDA and 93.9\% on Dream (AUROC 0.987 and 0.939; chance is 0.5).

\paragraph{The perturbation has no preferred direction.}
A zero-mean perturbation turns a near tie into a coin toss. Among commitments with a margin below 0.1, 47\% flip on LLaDA and 51\% on Dream.

\paragraph{The Gaussian curve matches the observed flip rates.}
With its single scale fitted to the data, the Gaussian curve differs from the observed flip rate by 0.1 percentage points on average across margins on LLaDA and 0.8 on Dream. The default scale $s=\tilde m$, fixed before the quantizer is known, overestimates the absolute probabilities; this matters little, since~(\ref{eq:prior}) keeps only the order of positions (last check below).\looseness=-1

\paragraph{Treating token and placement flips as independent is a close approximation.}
Both depend on the same confidences and are slightly correlated (about 0.1), but they rarely occur together: the predicted share of forced commitments that fail in either way is 37.8\% against 39.4\% observed on LLaDA, and 39.8\% against 39.2\% on Dream.

\paragraph{The prior does not depend on these modeling choices.}
Rebuilding the prior with the Gaussian curves replaced by flip rates measured directly from the data orders the 256 window positions almost exactly as ours (Spearman correlation 0.994 on LLaDA and 0.997 on Dream).

\paragraph{The reach.}
Its square root rests on approximately independent perturbations across later reads, a premise verified in \cref{app:proof}, where a logarithm in its place also gives a nearly identical prior. Reach alone improves on uniform weights by 1.1 (LLaDA) and 1.4 (Dream) average points, and the shuffled control of \cref{tab:ablation} shows that such gains require the weights to sit at the right positions.

\section{Drift-Guided Commitment (Polestar)}
\label{app:polestar}

We implemented the drift-guided commitment of Polestar \citep{lee2026polestar} in our Fast-dLLM pipeline (dual cache, $\tau=0.9$), so that the quantized checkpoints and the same-state measurement are those of the main paper, with its default settings: a position below the threshold is also committed when its drift jumps sharply relative to its recent history, and if no position qualifies, the most confident one is forced. At full precision, this raises the tokens committed per forward pass 1.22-fold, matching the 1.21-fold gain reported by Polestar. As argued in \cref{sec:decal}, a cache approximates attention, not the commitment rule, so we keep Fast-dLLM's dual cache.

\paragraph{Forced commitments remain the main source of errors.}
Forced commitments make up 11.6\% of commitments but carry 80.9\% of the permanent token flips; together with drift-admitted commitments, the decisions that bypass the threshold carry about 97\% of them, as under Fast-dLLM (\cref{tab:polestar-flips}).

\begin{table}[H]
\centering\small
\caption{Share of commitments and of permanent token flips by type, measured as in \cref{fig:obs}b (W4A4 FlatQuant, 400 GSM8K questions).}
\label{tab:polestar-flips}
\begin{tabular}{lcccc}
\toprule
 & \multicolumn{2}{c}{Drift-guided commitment} & \multicolumn{2}{c}{Fast-dLLM (\cref{fig:obs}b)} \\
\cmidrule(lr){2-3}\cmidrule(lr){4-5}
Commitment & commitments & token flips & commitments & token flips \\
\midrule
Above the threshold & 72.0\% & 2.6\% & 83.5\% & 2.9\% \\
Admitted by drift & 16.5\% & 16.6\% & -- & -- \\
Forced & 11.6\% & 80.9\% & 16.5\% & 97.1\% \\
\bottomrule
\end{tabular}
\end{table}

\paragraph{The parallelism tax persists, and \method{} keeps its gain.}
QuaRot needs 15.0\% more forward passes than FP16 (12.9\% under Fast-dLLM), because the drift test also depends on confidence, which quantization deflates; the FlatQuant-based models pay a small tax, as in \cref{fig:obs}c (\cref{tab:polestar-nfe}). At this speed, \method{} improves GSM8K accuracy over FlatQuant by 1.1 points (75.76 against 74.68, averaged over the three calibration runs of \cref{tab:main}), close to its 1.2-point gain under Fast-dLLM.

\begin{table}[H]
\centering\small
\caption{Forward passes per response on the full GSM8K test set under drift-guided commitment. FlatQuant and \method{}: mean of the three calibration runs of \cref{tab:main}.}
\label{tab:polestar-nfe}
\begin{tabular}{lcc}
\toprule
Model & Forward passes per response & vs.\ FP16 \\
\midrule
FP16 & 70.0 & -- \\
QuaRot & 80.4 & $+15.0\%$ \\
FlatQuant & 71.3 & $+1.9\%$ \\
\method{} & 71.1 & $+1.6\%$ \\
\bottomrule
\end{tabular}
\end{table}

\paragraph{The commit temperature complements drift-guided commitment.}
Drift-guided commitment speeds up both FP16 and W4A4 but does not remove the gap between them; the temperature, calibrated by~(\ref{eq:tmatch}) from forward-pass counts, does. \Cref{tab:polestar-temp} compares it with the other way to reach FP16's speed, a more aggressive drift test (smaller $\alpha$, within the range studied by Polestar). The temperature reaches FP16's speed without a significant accuracy change, whereas the aggressive drift test loses 11.4 points at the same speed. In the ladder used to set $T_{\mathrm{match}}$, forward passes fall monotonically as $T$ is lowered, from 79.7 to 66.7 per response on the calibration prompts as $T$ goes from 1 to 0.84.\looseness=-1

\begin{table}[H]
\centering\small
\caption{Bringing W4A4 QuaRot to the speed of FP16 (70.0 forward passes) under drift-guided commitment. Accuracy change: GSM8K, in points, relative to the default setting; n.s.: not statistically significant (paired test over the 1,319 questions). $\downarrow$/$\uparrow$: lower/higher is better.}
\label{tab:polestar-temp}
\begin{tabular}{lcc}
\toprule
W4A4 QuaRot setting & Forward passes $\downarrow$ & Accuracy change $\uparrow$ \\
\midrule
Default ($\alpha=10$, $T=1$) & 80.4 & -- \\
Commit temperature ($T=0.858$) & 69.1 & $-1.9$ (n.s.) \\
More aggressive drift test ($\alpha=5$) & 75.1 & $-3.0$ \\
More aggressive drift test ($\alpha=2.5$) & 69.4 & $-11.4$ \\
\bottomrule
\end{tabular}
\end{table}

\section{A Block-Diffusion Model: Fast-dLLM v2}
\label{app:v2}

Fast-dLLM v2 \citep{wu2026fastdllmv2} is a native block-diffusion model adapted from Qwen2.5-Instruct \citep{yang2024qwen25}, at 1.5B and 7B parameters. Among current block-diffusion models, it gives the cleanest test of our setting: its layers are dense like those of LLaDA and Dream, its 7B model is adapted, like Dream-7B, from Qwen2.5-7B (here its instruction-tuned version), and its decoder is the block-diffusion successor of Fast-dLLM, so block decoding is the main change. LLaDA2.0 \citep{bie2025llada2} uses mixture-of-experts layers at 16B and 100B parameters and would change architecture and scale together with the decoding scheme. Fast-dLLM v2 decodes blocks of 32 tokens from left to right with exact caching of completed blocks, and commits inside each block, in sub-blocks of 8 tokens, with the threshold rule~(\ref{eq:rule}) at $\tau=0.9$. \method{} is applied with its prior collected under this decoder; since v2-7B is adapted from the same base model as Dream, it uses Dream's setting $(\epsilon,c)=(1,0.15)$.\looseness=-1

\paragraph{Forced commitments and the parallelism tax carry over.}
Under W4A4, 10--15\% of the commitments in the returned output are forced, close to the 16.5\% on LLaDA (\cref{sec:obs1}). Quantization lowers the tokens written per forward pass and raises the share of forced commitments, and the resulting tax, 7.6\% on the 1.5B and 14.0\% on the 7B model, is well above FlatQuant's 0.8\% and 2.8\% on LLaDA and Dream (\cref{tab:v2-tax}).

\begin{table}[H]
\centering\small
\caption{Parallelism tax on Fast-dLLM v2 (200 GSM8K prompts, $\tau=0.9$, W4A4 FlatQuant). Forward passes are counted per returned token, so answer length does not enter.}
\label{tab:v2-tax}
\begin{tabular}{lccc}
\toprule
Model & Tokens per forward pass & Extra forward passes & Forced commitments \\
 & (FP16 $\to$ W4A4) & & (FP16 $\to$ W4A4) \\
\midrule
v2-1.5B & 2.10 $\to$ 1.95 & $+7.6\%$ & 12.6\% $\to$ 15.4\% \\
v2-7B & 2.58 $\to$ 2.27 & $+14.0\%$ & 8.2\% $\to$ 10.2\% \\
\bottomrule
\end{tabular}
\end{table}

\paragraph{\method{} improves W4A4 accuracy.}
On v2-7B, \method{} improves on FlatQuant by 1.1 points on GSM8K and 0.9 on MATH, closing about a fifth of the gap to full precision, in line with its gains in \cref{tab:main} (\cref{tab:v2-acc}).\looseness=-1

\begin{table}[H]
\centering\small
\caption{W4A4 accuracy (\%) on Fast-dLLM v2-7B ($\tau=0.9$; mean $\pm$ std over 3 calibration runs).}
\label{tab:v2-acc}
\begin{tabular}{lccc}
\toprule
Method & GSM8K & MATH & Avg. \\
\midrule
FP16 & 82.11 & 46.07 & 64.09 \\
FlatQuant & 77.94\,{\scriptsize$\pm$0.77} & 39.05\,{\scriptsize$\pm$0.64} & 58.49 \\
\method{} & \textbf{79.02}\,{\scriptsize$\pm$1.69} & \textbf{39.97}\,{\scriptsize$\pm$0.52} & \textbf{59.50} \\
\bottomrule
\end{tabular}
\end{table}

\section{Sensitivity to the Concentration \texorpdfstring{$c$}{c}}
\label{app:csens}

\begin{figure}[H]
  \centering
  \includegraphics[width=0.98\linewidth]{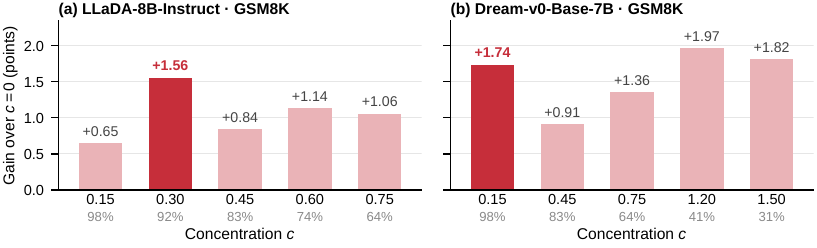}
  \caption{Sensitivity of \method{} to the concentration $c$: gain in W4A4 GSM8K test accuracy over uniform weights ($c=0$), mean of two calibration runs per value. Dark bars: values used in the paper. Gray percentages: effective sample size $K/(1+c^2)$ relative to $K$.}
  \label{fig:csens}
\end{figure}

The concentration $c$ is the only free parameter of the prior~(\ref{eq:prior}). It is the coefficient of variation of the window weights and sets how strongly the calibration budget is concentrated on the high-risk positions, whose ranking comes from the decoder and does not change with $c$. Concentration has a price: the effective sample size falls to $K/(1+c^2)$, and $c$ is bounded by $c_{\max}=\mathrm{CV}(r)/(1-r_{\min}/\bar r)$, at which the weight of the least risky position reaches zero (0.79 on LLaDA, 1.60 on Dream). For the main results (\cref{sec:exp-setup}), $c$ was set once per model on the GSM8K test set after a quick check of a few moderate values; both values keep most of the effective sample size (92\% on LLaDA, 98\% on Dream). To measure how much this choice matters, we vary $c$ from 0 to about 95\% of $c_{\max}$, keep everything else as in \cref{tab:ablation}, calibrate each value twice, and evaluate on the full GSM8K test set (\cref{fig:csens}).\looseness=-1

\paragraph{The gain requires enough concentration.}
At $c=0$ the prior has no effect, and \method{} reduces to uniform weights. On LLaDA, $c=0.15$ recovers only part of the gain (0.65 points); on Dream, most of it is already reached at $c=0.15$ (1.74 points).

\paragraph{Beyond that point, the gain is stable.}
From $c=0.30$ on LLaDA and $c=0.15$ on Dream up to near $c_{\max}$, every value tested improves on uniform weights by 0.8--1.6 (LLaDA) and 0.9--2.0 (Dream) points, within about one point of each other, even as the effective sample size falls to 64\% and 31\%. Once the budget is concentrated on the high-risk positions, further concentration adds little, as expected from the trade-off between focus and effective sample size (\cref{sec:decal}). The values used in the paper lie at the start of this plateau, and their gains are close to the GSM8K gains in \cref{tab:ablation} (1.74 points on both models), which come from separate calibration runs.

\end{document}